\documentclass[journal,twoside]{IEEEtran}
\usepackage{amsfonts,bm,booktabs,cite,graphicx,hyperref,multirow,textcomp,url}
\usepackage{algorithm}
\usepackage{subfigure}
\usepackage{algpseudocode}
\usepackage{amsmath}
\usepackage{mathrsfs}
\usepackage{float}
\usepackage{color}
\usepackage{booktabs}
\usepackage{bbding}
\usepackage[flushleft]{threeparttable}

\makeatletter 
\renewcommand{\@thesubfigure}{\hskip\subfiglabelskip}
 \makeatother
\IEEEpubid{\begin{minipage}{\textwidth}\ \centering
		Copyright \copyright 2024 IEEE.  Personal use of this material is permitted.  Permission from IEEE must be obtained for all other uses, in any current or future media, including reprinting/republishing this material for advertising or promotional purposes, creating new collective works, for resale or redistribution to servers or lists, or reuse of any copyrighted component of this work in other works.
\end{minipage}}

\begin{document}
\title{Bi-Directional Deep Contextual Video Compression}
\author{
	Xihua Sheng, \IEEEmembership{Member, IEEE},
	Li Li, \IEEEmembership{Member, IEEE},
	Dong Liu, \IEEEmembership{Senior Member, IEEE}\\
	Shiqi Wang, \IEEEmembership{Senior Member, IEEE}\\
\thanks{
Date of current version \today.\par 
This work was supported in part by the Natural Science Foundation of China under Grants 62171429/62021001 and in part by RGC General Research Fund 11203220/11200323. It was also supported by the GPU cluster built by MCC Lab of Information Science and Technology Institution, USTC.\par

X. Sheng, L. Li, and D. Liu are with the MoE Key Laboratory of Brain-inspired Intelligent Perception and Cognition, University of Science and Technology of China, Hefei 230027, China. L. Li is also with the Institute of Artificial Intelligence, Hefei Comprehensive National Science Center (e-mail: xhsheng@mail.ustc.edu.cn, lil1@ustc.edu.cn, dongeliu@ustc.edu.cn). \par
S. Wang is with the Department of Computer Science, City University of Hong Kong, Hong Kong, China (e-mail: shiqwang@cityu.edu.hk).\par
Corresponding author: Li Li.\par
}
}

\markboth{IEEE Transactions on Multimedia}{Bi-Directional Deep Contextual Video Compression}

\maketitle
\begin{abstract}
Deep video compression has made impressive process in recent years, with the majority of advancements concentrated on P-frame coding. Although efforts to enhance B-frame coding are ongoing, their compression performance is still far behind that of traditional bi-directional video codecs. In this paper, we introduce a bi-directional deep contextual video compression scheme tailored for B-frames, termed DCVC-B, to improve the compression performance of deep B-frame coding. Our scheme mainly has three key innovations. First, we develop a bi-directional motion difference context propagation method for effective motion difference coding, which significantly reduces the bit cost of bi-directional motions. Second, we propose a bi-directional contextual compression model and a corresponding bi-directional temporal entropy model, to make better use of the multi-scale temporal contexts. Third, we propose a hierarchical quality structure-based training strategy, leading to an effective bit allocation across large groups of pictures (GOP). Experimental results show that our DCVC-B achieves an average reduction of 26.6\% in BD-Rate compared to the reference software for H.265/HEVC under random access conditions. Remarkably, it surpasses the performance of the H.266/VVC reference software on certain test datasets under the same configuration. We anticipate our work can provide valuable insights and bring up deep B-frame coding to the next level. 
\end{abstract}
\IEEEpeerreviewmaketitle

\section{Introduction}
With the popularity of various video applications, video data has accounted for most of the global Internet traffic. Massive video data brings huge costs to video transmission and video storage. Therefore, it is urgent to compress videos efficiently.\par

In the past decades, several traditional video coding standards have been developed such as H.264/AVC~\cite{wiegand2003overview}, H.265/HEVC~\cite{sullivan2012overview}, and H.266/VVC~\cite{bross2021overview}, which greatly improve the video compression performance. In terms of different kinds of video applications, various coding configurations are designed. Among them, low delay and random access are two typical configurations. The low delay configuration is used in applications where minimizing latency is crucial, such as video conferencing, live streaming, and video gaming. In this configuration, only the reference frames that occurred in the past of the current frame (P-frame) can be used. The random access configuration is used in applications where there is no need to decode the entire bitstream from the beginning, such as video-on-demand, broadcasting, and content delivery networks. In this configuration, the reference frames that occurred in the past and future of the current frame (B-frame) can be used. Since bi-directional temporal information can be utilized, the random access configuration designed for B-frame coding can achieve higher compression performance than the low delay configuration designed for P-frame coding. Taking the reference software of H.265/HEVC standard as an example, its compression performance under the random access configuration is 30\% higher than that under the low delay configuration.\par
Although traditional video coding standards have achieved great success, it is more and more challenging to achieve large compression performance improvements under limited coding complexity increases. With the development of deep neural networks, deep video compression schemes have been proposed to break through the bottleneck of compression performance. Existing deep video compression schemes can be divided into two classes, including P-frame coding schemes and B-frame coding schemes. Among them, P-frame coding schemes develop faster. Especially after the emergence of advanced motion-compensated predictions~\cite{hu2021fvc,agustsson2020scale,sheng2024spatial,shi2022alphavc},  conditional coding~\cite{li2021deep,li2022hybrid,li2023neural,sheng2022temporal}, and efficient training strategies~\cite{li2024neural,sheng2022temporal,lu2020content}, their compression performance even exceeds that of the reference software of H.266/VVC under the low delay configuration. However, the compression performance of B-frame coding schemes is still much lower than that of traditional video codecs. The best deep B-frame coding schemes~\cite{chen2023b,yang2024ucvc} are only comparable to the reference software of H.265/HEVC under the random access configuration.\par
\IEEEpubidadjcol

There are three main reasons for the low compression performance of existing deep B-frame coding schemes. The first reason is that compressing bi-directional motion vectors needs more motion coding costs. Although some schemes~\cite{yang2020learning,yilmaz2021end} proposed motion difference coding methods, the motion redundancy is not fully reduced and the motion bitrate increment is still unaffordable. The second reason is that the temporal predictions cannot be fully used. Although some schemes~\cite{chen2023b,yang2024ucvc} have applied conditional coding to utilize feature-based temporal predictions, the temporal correlation is still utilized sufficiently by different coding modules. The third reason is that the training strategy is not efficient. Most training strategies cannot help build a hierarchical quality structure across a large GOP, resulting in inappropriate bit allocation.\par

In terms of the three limitations, we propose a bi-directional deep contextual video compression scheme tailored for B-frames, termed DCVC-B.  The main contributions of our proposed scheme are summarized as follows.
\begin{itemize}
    \item We propose a bi-directional motion difference context propagation method for effective motion difference coding, which significantly reduces the motion coding costs.

    \item We propose a bi-directional context compression model and a corresponding bi-directional temporal entropy model to make better use of multi-scale temporal contexts. 

    \item We propose a hierarchical quality structure-based training strategy, which can achieve a better bit allocation across a large GOP.
\end{itemize}
Experimental results show that in terms of PSNR, our proposed DCVC-B scheme outperforms the reference software of H.265/HEVC under the random access configuration by a large margin (26.6\% on average). It even outperforms the reference software of H.266/VVC under the random access configuration on some testing datasets.\par

The remainder of this paper is organized as follows.  Section~\ref{sec:related_work} gives a review of related work about deep video compression for P-frame and B-frame. Section~\ref{sec:overview} introduces a brief overview of our proposed DCVC-B scheme. Section~\ref{sec:methodology} describes our proposed methods in detail. Section~\ref{sec:experiments} presents the experimental results and ablation studies. Section~\ref{sec:conclusion} gives a conclusion of this paper.

\section{Related Work}\label{sec:related_work}
\subsection{Deep Video Compression for P-Frame}\label{sec:related_work_p}
A P-frame can only refer to the frames that occur before it. Most deep video compression schemes focus on P-frame coding~\cite{jin2023learned,Habibian_2019_ICCV,wu2018video,rippel2019learned,DBLP:conf/cvpr/LuO0ZCG19,chen2019learning,lu2020end,liu2020learned,liu2020neural,liu2020conditional,agustsson2020scale,lin2020m,hu2021fvc,yilmaz2021end,Rippel_2021_ICCV,li2021deep,hu2022coarse,liu2022end,shi2022alphavc,DBLP:conf/nips/MentzerTMCHLA22,xiang2022mimt,lin2022dmvc,yang2022advancing,guo2023learning,guo2023enhanced,wang2023learned,ma2024uncertainty,kim2023c3,sheng2022temporal,li2024neural,tang2024offline,lu2024deep,sheng2024vnvc,du2024cgvc,sheng2024spatial}. Based on the first deep P-frame coding scheme---DVC~\cite{lu2020end},  subsequent schemes mainly focus on three aspects to improve compression performance, including (1) how to obtain more accurate temporal predictions with less motion coding costs, (2) how to make better use of temporal predictions to reduce temporal redundancy, and (3) how to design more efficient training strategies. \par
For the first aspect, Agustsson et al.~\cite{agustsson2020scale} proposed a scale-space flow to blur regions with inaccurate predictions to reduce prediction errors. Lin et al.~\cite{lin2020m} proposed a multiple frame-based motion vector prediction method and a multiple frame-based motion compensation method to reduce the motion coding cost and improve the prediction accuracy. Tang et al.~\cite{tang2024offline} proposed an offline and online optical flow enhancement method, which uses the motion vectors estimated by VTM as labels to optimize the optical flow estimation network and online update the latent representations of motion encoder-decoder according to the video contents. With the proposed method, more accurate temporal prediction can be obtained with less motion coding costs. Except for flow-based temporal prediction, Hu et al.~\cite{hu2021fvc} proposed a feature-based video compression framework. With the introduced deformable convolution, the accuracy of motion estimation and the effectiveness of motion compensation can be increased.  \par

For the second aspect, Li~et al.~\cite{li2021deep} proposed a deep contextual video compression scheme, termed DCVC, which replaces the residual coding paradigm with a conditional coding paradigm. Instead of relying on the subtraction operation to reduce temporal redundancy, DCVC regards the temporal prediction as a condition so that the codec can learn how to make full use of the temporal prediction automatically. To further utilize temporal predictions, Sheng~et al.~\cite{sheng2022temporal}  proposed DCVC-TCM, which not only uses temporal predictions at the entrance of the contextual encoder but also feeds the multi-scale temporal predictions into its intermediate locations. In addition, they also proposed to learn a temporal prior from the multi-scale temporal predictions to utilize temporal predictions in the entropy model. Based on DCVC-TCM, Li~\cite{li2022hybrid} proposed DCVC-HEM, which introduced a latent prior into the temporal entropy model to further utilize temporal correlation. \par

For the third aspect, Lin et al.~\cite{lin2020m} and Sheng et al.~\cite{sheng2022temporal} proposed a step-by-step training strategy. First, the motion-dependent modules are optimized, then the residual/context-dependent modules are optimized, and finally, all the modules are jointly optimized. The step-by-step training strategy is beneficial to the optimization of a deep video codec with multiple coding modules. To mitigate the error propagation, Lu et al.~\cite{lu2020content} and Sheng et al.~\cite{sheng2022temporal} proposed a multi-frame cascaded fine-tuning strategy. In the fine-tuning process, the losses of multiple frames are averaged and used to optimize the codec. Li et al.~\cite{li2023neural} further proposed to assign a coefficient that varies periodically over the frame index to the Lagrangian multiplier in the rate-distortion (R-D) loss function to periodically increase the quality of video frames so that the error propagation can be further reduced. 

\par

Relying on advanced video coding technologies and training strategies, the compression performance of deep P-frame coding schemes has outperformed the reference software of H.266/VVC~\cite{bross2021overview} in terms of PSNR.

\subsection{Deep Video Compression for B-Frame}~
A B-frame can refer to frames that occur before and after it. Existing deep B-frame coding schemes can be divided into two classes, including the schemes without motion coding and the schemes with motion coding. \par
In the first class, most of the schemes use video interpolation methods to get temporal predictions~\cite{wu2018video, djelouah2019neural, alexandre2023hierarchical, xu2024ibvc}. For example, Wu~\cite{wu2018video} proposed a contextual video interpolation network to obtain a predicted frame of the current frame using the multi-scale contexts of forward and backward reference frames. Then, they compress the residual between the current and predicted frames. Under the test condition of intra period 12 and GOP size 12, their compression performance is comparable to that of x264 with \emph{fast} preset, the industrial software of H.264/AVC~\cite{wiegand2003overview}. Similar to~\cite{wu2018video}, Djelouah~et al.~\cite{djelouah2019neural} proposed to reduce temporal redundancy by transforming the current and interpolated frames to latent codes and calculating the residual of latent codes. Under the same test condition as~\cite{wu2018video}, they surpassed x264 with \emph{fast} preset. Alexandre et al.~\cite{alexandre2023hierarchical} proposed a two-layer B-frame coding architecture. At the base layer, the downsampled current frame is compressed by an image codec with the condition of a low-resolution interpolated frame. The reconstructed low-resolution frame is merged with the high-resolution predicted frames to generate a high-quality image as a condition for the enhancement layer. Under the condition of intra period 32 and GOP32 (the difficult setting of the random access configuration of the reference software of H.266/VVC), they outperformed x265 with \emph{very slow} preset, the industrial software of H.265/HEVC~\cite{sullivan2012overview}. 

In the second class, Yang et al.~\cite{yang2020learning} proposed to estimate and compress the motion vectors between the current frame and bi-directional reference frames. After reconstructing the motion vectors, they perform bi-directional motion compensation to obtain a predicted frame. To achieve a hierarchical quality structure, they transmitted different target quality and bitrate settings to the decoder and generated hierarchical quality weights from these settings. Under the test condition of intra period 10 and GOP size 10, they achieved a better compression performance than x265 with \emph{very fast} preset. Concurrently with~\cite{yang2020learning}, Yilmaz et al.~\cite{yilmaz2020end} proposed a U-shaped mask generation network to merge the bi-directional predicted frames better. Based on this work, their extended work~\cite{yilmaz2021end} further proposed novel tools such as motion vector subsampling and surpassed x265 with \emph{veryslow} preset. Chen~\cite{chen2023b} proposed a normalizing flows-based B-frame coding scheme, which can dynamically adapt the feature distributions according to the B-frame type and allow better flexibility in specifying the GOP structure. Under the condition of intra period 32 and GOP size 16 (the difficult setting of the random access configuration of the reference software of H.265/HEVC), they achieved comparable compression results to the reference software of H.265/HEVC~\cite{sullivan2012overview} under the random access configuration in terms of PSNR. Feng et al.~\cite{feng2021versatile} proposed a versatile P-frame and B-frame coding scheme, which used a voxel flow and a trilinear warping operation to obtain a predicted frame from multiple uni-directional or bi-directional reference frames. Under the condition of intra period 12 and GOP 12, they outperformed the reference software of H.266/VVC~\cite{bross2021overview} in terms of MS-SSIM. Similarly, Yang~et al.~\cite{yang2024ucvc} also proposed a unified scheme for P-frame and B-frame but they shifted the residual coding paradigm to conditional coding paradigm~\cite{li2021deep,sheng2022temporal}. Under the condition of intra period 32 and GOP size 32, they obtained a compression performance comparable to DCVC-HEM in terms of PSNR.\par

In summary, we can observe that the compression performance of existing deep B-frame coding schemes under standard random access configurations (intra period 32, GOP size 16 or intra period 32, GOP size 32) is still far from that of reference software of traditional video coding standards in terms of PSNR. Therefore, in this work, we focus on improving the compression performance of deep B-frame coding. Similar to existing successful deep P-frame coding schemes, we improve performance from three aspects, including (1) reducing the bi-directional motion coding costs, (2) making better use of bi-directional temporal predictions, and (3) designing a more efficient training strategy.
\begin{figure*}[t]
  \centering
   \includegraphics[width=\linewidth]{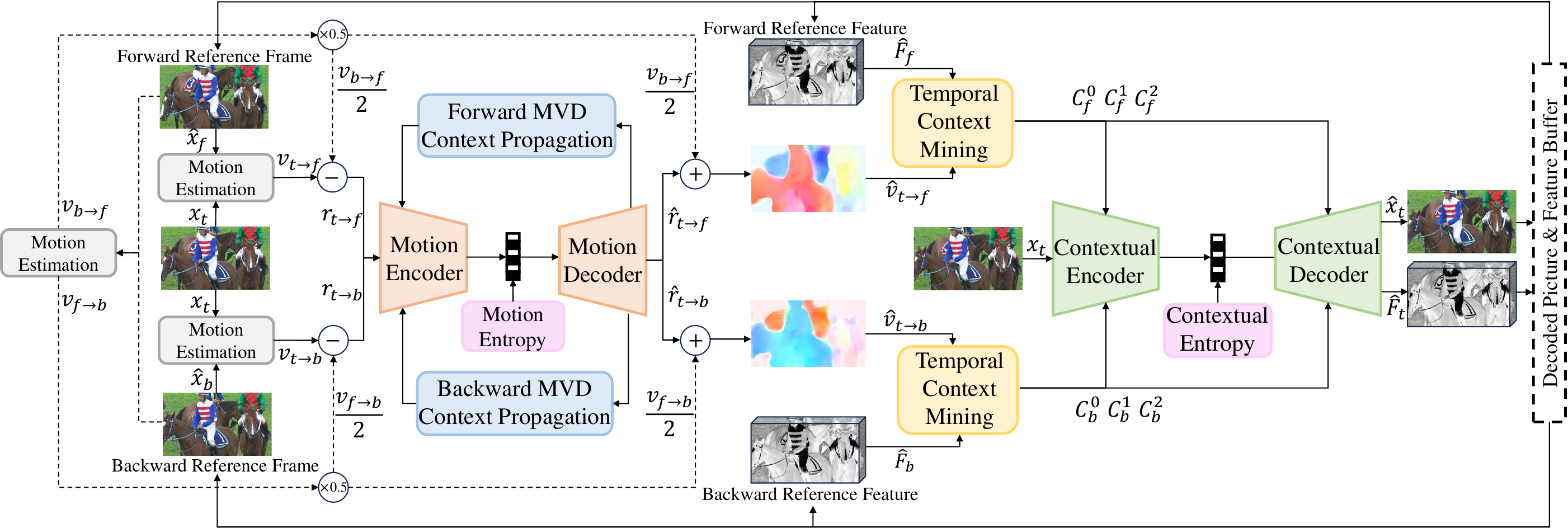}
      \caption{Overview of our proposed bi-directional deep contextual video compression scheme---DCVC-B. The motion estimation module estimates the bi-directional motion vectors ($v_{t\rightarrow f}$, $v_{t\rightarrow b}$) between the current frame $x_t$ and bi-directional reference frames ($\hat{x}_{f}$, $\hat{x}_{b}$) and also estimates the motion vector predictions ($v_{b\rightarrow f}$, $v_{f\rightarrow b}$) between ($\hat{x}_{f}$, $\hat{x}_{b}$). Then the motion vector differences (MVD) ($r_{t\rightarrow f}$, $r_{t\rightarrow b}$) between ($v_{t\rightarrow f}$, $v_{t\rightarrow b}$) and their predictions ($\frac{v_{b\rightarrow f}}{2}$, $\frac{v_{f\rightarrow b}}{2}$) are jointly compressed and decompressed by a motion encoder-decoder with our proposed bi-directional motion difference context propagation method. The reconstructed motion vectors ($\hat{v}_{t\rightarrow f}$, $\hat{v}_{t\rightarrow b}$) are used to perform bi-directional temporal context mining over the bi-directional reference features ($\hat{F}_{f}$, $\hat{F}_{b}$). The predicted bi-directional multi-scale temporal contexts ($C_f^0$, $C_f^1$, $C_f^2$), ($C_b^0$, $C_b^1$, $C_b^2$) are fed into a contextual encoder-decoder to help compress and decompress the current frame $x_t$. Before obtaining the reconstructed frame $\hat{x}_t$, we regard an intermediate feature $\hat{F}_t$ of the contextual decoder as the propagated reference feature.}
   \label{fig:framework}
\end{figure*}
\begin{figure*}[t]
  \centering
   \includegraphics[width=\linewidth]{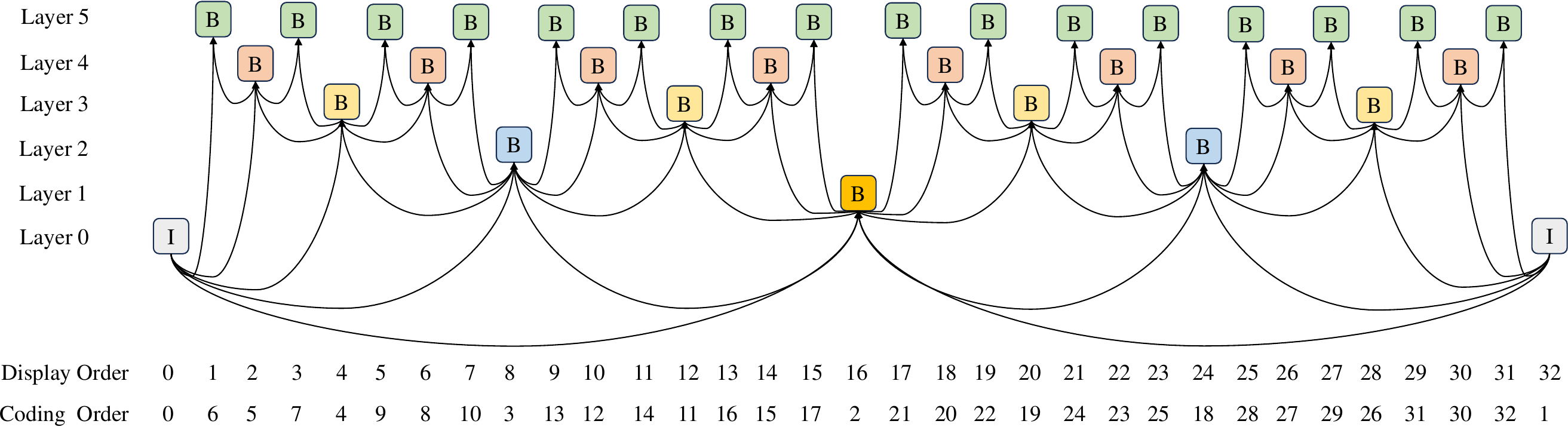}
      \caption{Structure of the group of pictures (GOP) of our proposed DCVC-B scheme. Following the default random access configuration of reference software of H.266/VVC~\cite{bross2021overview}, we set the intra period and GOP size to 32. There are six temporal layers within a GOP. We assign different quality coefficients for the B-frames in different temporal layers to achieve a hierarchical quality structure.}
   \label{fig:GOP_structure}
\end{figure*}

\section{Overview}\label{sec:overview}
We first summarize the overview architecture of our proposed bi-directional deep video compression scheme. 
\subsubsection{GOP Structure}
We design a hierarchical GOP structure for our proposed DCVC-B scheme as illustrated in Fig.~\ref{fig:GOP_structure}. Following the default random access configuration of the reference software of H.266/VVC~\cite{bross2021overview} standard---VTM~\cite{VTM}, we set both the intra period and the GOP size to 32. There are six temporal layers in a GOP. Among these temporal layers, Layer 0 consists of I-frames and the other layers consist of B-frames. We assign different hierarchical quality coefficients for the B-frames in different temporal layers when training the B-frame coding scheme. The details of the hierarchical quality structure-based training strategy will be described in Section~\ref{training_strategy}.

\subsubsection{Bi-Directional Motion Estimation}
We use a pre-trained SpyNet~\cite{ranjan2017optical} to estimate the bi-directional motion vectors ($v_{t\rightarrow f}$, $v_{t\rightarrow b}$) between the current frame $x_t$ and the forward and backward reference frames ($\hat{x}_{f}$, $\hat{x}_{b}$). Following~\cite{yang2020learning,yilmaz2021end}, we also estimate the bi-directional motion vectors ($v_{b\rightarrow f}$, $v_{f\rightarrow b}$) between ($\hat{x}_{f}$, $\hat{x}_{b}$) as the motion vector predictions of ($v_{t\rightarrow f}$, $v_{t\rightarrow b}$).

\subsubsection{Bi-Directional Motion Compression}
We design a variable-bitrate motion encoder-decoder as shown in Fig.~\ref{fig:framework}. To reduce the motion coding costs, we calculate the motion vector differences (MVDs) between the input motion vectors ($v_{t\rightarrow f}$, $v_{t\rightarrow b}$) and their predictions ($\frac{v_{b\rightarrow f}}{2}$, $\frac{v_{f\rightarrow b}}{2}$) following~\cite{yang2020learning,yilmaz2021end}. Then we perform channel-wise concatenation to the motion vector differences ($r_{t\rightarrow f}$, $r_{t\rightarrow b}$) and jointly compress them into a compact latent representation $m_t$ with the size of $H/16 \times W/16 \times C_m$. After quantization, we use an arithmetic encoder to signal the quantized motion representation $\hat{m}_t$ into a bitstream and transmit it to the decoder. At the decoder side, we decompress $\hat{m}_t$ using an arithmetic decoder and inversely transform it back to the reconstructed motion vector differences ($\hat{r}_{t\rightarrow f}$, $\hat{r}_{t\rightarrow b}$). The reconstructed motion vector differences are added to the corresponding predictions ($\frac{v_{b\rightarrow f}}{2}$, $\frac{v_{f\rightarrow b}}{2}$) to obtain the reconstructed bi-directional motion vectors ($\hat{v}_{t\rightarrow f}$, $\hat{v}_{t\rightarrow b}$). To further improve the motion compression efficiency, we propose a bi-directional motion difference context propagation method, which will be described in detail in Section~\ref{sec:MCP}.

\subsubsection{Bi-Directional Temporal Context Mining}
To make full use of temporal information, as shown in Fig.~\ref{fig:framework}, we propose to propagate bi-directional temporal information through forward and backward reference features ($\hat{F}_{f}$, $\hat{F}_{b}$) instead of pixel-domain reference frames ($\hat{x}_{f}$, $\hat{x}_{b}$). To generate more accurate temporal predictions, we shift the temporal context mining from P-frame coding~\cite{sheng2022temporal} to B-frame coding and predict multi-scale bi-directional temporal contexts ($C_{t}^{f,l}$, $C_{t}^{b,l}$) from the bi-directional reference features ($\hat{F}_{f}$, $\hat{F}_{b}$). The details can be found in Section~\ref{sec:TCM}.
\subsubsection{Bi-Directional Contextual Compression}
After learning the bi-directional multi-scale temporal contexts ($C_{t}^{f,l}$, $C_{t}^{b,l}$), we regard them as conditions and feed them into a variable-bitrate contextual encoder-decoder. The input frame $x_t$ is compressed into a compact latent representation $y_t$ with the size of $H/16 \times W/16 \times C_y$. 
In the compressing process, we propose to feed the multi-scale bi-directional temporal contexts ($C_{t}^{f,l}$, $C_{t}^{b,l}$) into the contextual encoder to make better use temporal predictions. Then $y_t$ is quantized and its bitstream signaled by the arithmetic encoder is transmitted to the decoder.  At the decoder side, we reconstruct the bitstream back to $\hat{y}_t$ using the arithmetic decoder and decompress $\hat{y}_t$ to the reconstructed frame $\hat{x}_t$ using the contextual decoder. In the decompressing process, we also feed the multi-scale bi-directional temporal contexts ($C_{t}^{f,l}$, $C_{t}^{b,l}$) into the contextual decoder to complement the temporal information. We will describe the details of bi-directional contextual compression in Section~\ref{sec:contextual_ED}.

\subsubsection{Entropy Model}
We use the factorized entropy model~\cite{DBLP:conf/iclr/BalleLS17} for hyperprior and the Laplace distribution~\cite{DBLP:conf/iclr/BalleMSHJ18} to model the motion and contextual compact latent representations $\hat{m}_t$ and $\hat{y}_t$.  We combine the hyperprior, the spatial prior generated by the quadtree partition-based spatial entropy model~\cite{sheng2024spatial,li2023neural}, and the temporal prior generated by our proposed bi-directional temporal entropy model together to estimate the mean and scale of the Laplace distributions of $\hat{m}_t$ and $\hat{y}_t$.  The details of the bi-directional temporal entropy model will be introduced in Section~\ref{sec:TEM}.
\begin{figure*}[t]
  \centering
   \includegraphics[width=\linewidth]{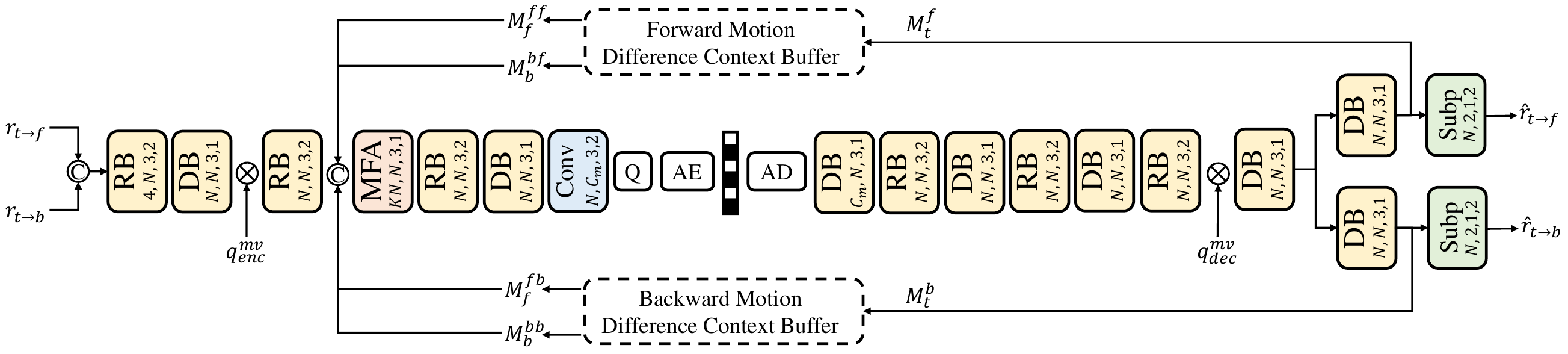}
      \caption{Architecture of the motion encoder-decoder with our proposed bi-directional motion difference context propagation method. ``RB" refers to residual block. ``DB" refers to depth block~\cite{li2023neural}. ``Subp" refers to the subpixel layer~\cite{shi2016real}. ``MFA" refers to the motion feature adaptor.}
   \label{fig:MED}
\end{figure*}
\begin{figure}[t]
  \centering
   \includegraphics[width=0.7\linewidth]{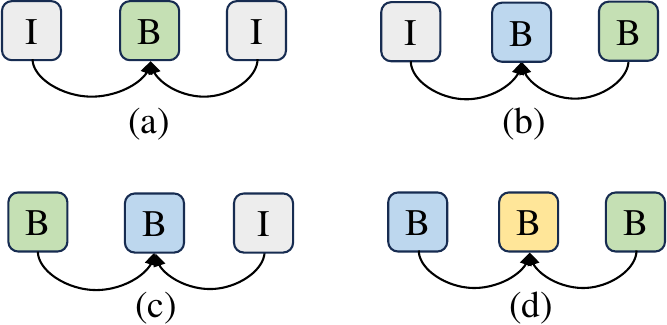}
      \caption{Different types of reference information propagation.}
   \label{fig:reference_propagation}
\end{figure}

\section{Methodology}\label{sec:methodology}
\subsection{Bi-directional Motion Difference Context Propagation}~\label{sec:MCP}
Since bi-directional motion vectors need to be compressed, the motion coding costs of deep B-frame coding increase significantly. Although we follow~\cite{yang2020learning,yilmaz2021end} to perform subtraction operations between the input motion vectors and their predictions as shown in Fig.~\ref{fig:framework}, the motion bitrate increment is still not neglectable. Considering the forward and backward reference frames ($\hat{x}_f$, $\hat{x}_b$) have their own motion vector differences ($\hat{r}_{f\rightarrow ff}$, $\hat{r}_{f\rightarrow fb}$), ($\hat{r}_{b\rightarrow bf}$, $\hat{r}_{b\rightarrow bb}$), where ($ff$, $fb$) are the bi-directional reference frame indexes of $\hat{x}_f$, and ($bf$, $bb$) are the bi-directional reference frame indexes of $\hat{x}_b$,  we can utilize the temporal information of their motion vector differences to further reduce the coding costs of the current motion vector differences ($r_{t\rightarrow f}$, $r_{t\rightarrow b}$) of $x_t$. Therefore, we propose a bi-directional motion difference context propagation method as illustrated in Fig.~\ref{fig:MED}.\par

When compressing the current bi-directional motion vector differences ($r_{t\rightarrow f}$, $r_{t\rightarrow b}$), we regard the feature-based forward and backward motion difference contexts ($M_f^{ff}$, $M_b^{bf}$), ($M_f^{fb}$, $M_b^{bb}$) as conditions and feed them into the motion encoder. 
For different types of reference frames, we design different motion feature adaptors (MFAs) implemented by depth blocks~\cite{li2023neural} to fuse their motion difference contexts. When both the forward and backward reference frames are I-frames, as shown in Fig.~\ref{fig:reference_propagation}(a), the $MFA_0$ is used. No motion difference context is fused.
\begin{equation}
    MF_{out}= MFA_0(MF_{in}).
\end{equation}
When the forward reference frame is I-frames and the backward reference frame is B-frame, as shown in Fig.~\ref{fig:reference_propagation}(b), the $MFA_1$ is used. The motion difference contexts ($M_b^{bf}$, $M_b^{bb}$) of the backward reference frame are fused.
\begin{equation}
    MF_{out}= MFA_1(concat(MF_{in}, M_b^{bf}, M_b^{bb})).
\end{equation}
When the forward reference frame is B-frames and the backward reference frame is I-frame, as shown in Fig.~\ref{fig:reference_propagation}(c), the $MFA_2$ is used. The motion difference contexts ($M_f^{ff}$, $M_f^{fb}$) of the forward reference frame are fused.
\begin{equation}
    MF_{out}= MFA_2(concat(MF_{in}, M_f^{ff}, M_f^{fb})).
\end{equation}
When the forward and backward reference frames are both B-frames, as shown in Fig.~\ref{fig:reference_propagation}(d), the $MFA_3$ is used. The motion difference contexts ($M_f^{ff}$, $M_b^{fb}$), ($M_b^{bf}$, $M_b^{bb}$)  of the forward and backward reference frames are fused.
\begin{equation}
    MF_{out}= MFA_3(concat(MF_{in}, M_f^{ff}, M_f^{fb}, M_b^{bf}, M_b^{bb})).
\end{equation}
$MF_{in}$ is the input feature of the motion feature adaptor and $MF_{out}$ is its output feature. The operation $concat$ refers to channel-wise concatenation. Before obtaining the reconstructed bi-directional motion vector difference ($\hat{r}_{t\rightarrow f}$, $\hat{r}_{t\rightarrow b}$), we regard the input features ($M_t^f$, $M_t^b$) of the last sub-pixel layers~\cite{shi2016real} as the motion difference contexts of the current frame. To make one model support variable bitrates, we insert two learnable quantization steps ($q_{dec}^{mv}$,$q_{enc}^{mv}$)~\cite{sheng2024spatial,li2023neural} into the motion encoder and decoder, respectively. 
\begin{figure}[t]
  \centering
   \includegraphics[width=0.8\linewidth]{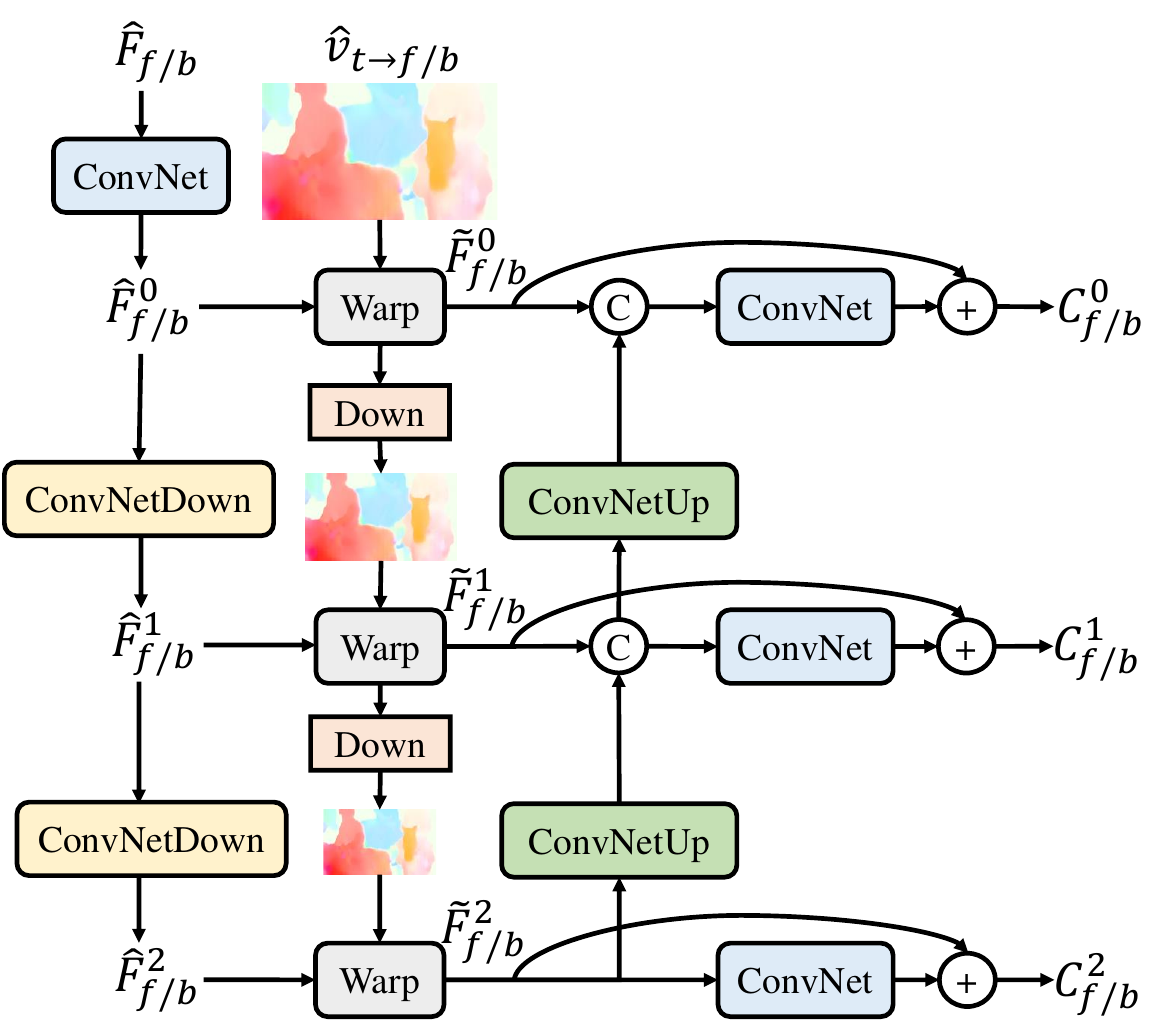}
      \caption{Architecture of the bi-directional temporal context mining module. The ``Convnet" is implemented by a convolutional layer and a residual block. The ``ConNetDown" is implemented by a convolutional layer with stride 2 and a residual block. The ``ConNetUP" is implemented by a subpixel layer and a residual block.}
   \label{fig:Bi_TCM}
\end{figure}

\begin{figure*}[t]
  \centering
   \includegraphics[width=\linewidth]{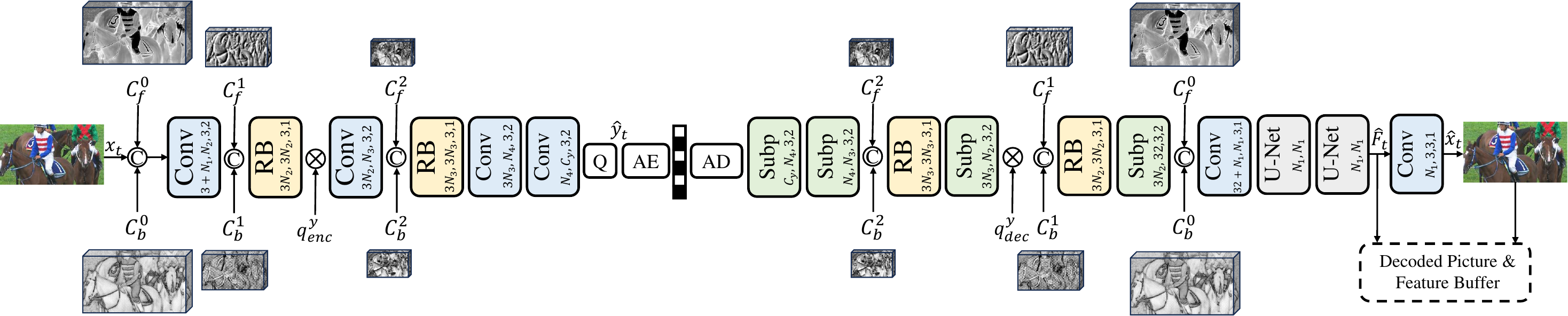}
      \caption{Architecture of the bi-directional contextual encoder-decoder. The multi-scale bi-directional temporal contexts are fed into the contextual encoder-decoder to reduce temporal redundancy.}
   \label{fig:CED}
\end{figure*}
\subsection{Bi-directional Temporal Context Mining}~\label{sec:TCM}
To generate more accurate temporal predictions, we propose to generate multi-scale feature-domain temporal predictions from bi-directional reference features ($\hat{F}_f$, $\hat{F}_b$). Specially, when the forward and backward reference frames are both B-frame, as shown in Fig.~\ref{fig:Bi_TCM}, we use a feature extractor (FE) module to obtain multi-scale bi-directional reference features ($\hat{F}_f^0$, $\hat{F}_f^1$, $\hat{F}_f^2$), ($\hat{F}_b^0$, $\hat{F}_b^1$, $\hat{F}_b^2$) from their reference features ($\hat{F}_f$, $\hat{F}_b$), respectively. 
The multi-scale reference features have the same resolution, half of the resolution, and a quarter of the resolution as ($\hat{F}_f$, $\hat{F}_b$), respectively. 
\begin{equation}
    \hat{F}_f^l= FE_f(\hat{F}_f), l=0,1,2.
\end{equation}
\begin{equation}
    \hat{F}_b^l= FE_b(\hat{F}_b), l=0,1,2.
\end{equation}
We also use the bilinear downsampling operation to obtain multi-scale bi-directional motion vectors ($\hat{v}_f^0$, $\hat{v}_f^1$, $\hat{v}_f^2$), ($\hat{v}_b^0$, $\hat{v}_b^1$, $\hat{v}_b^2$) from ($\hat{v}_f$, $\hat{v}_b$), respectively. We set ($\hat{v}_f^0$, $\hat{v}_b^0$) to ($\hat{v}_f$, $\hat{v}_b$). 
\begin{equation}
    \hat{v}_f^{l+1}= bilinear(\hat{v}_f^l)/2, l=0,1.
\end{equation}
\begin{equation}
    \hat{v}_b^{l+1}= bilinear(\hat{v}_b^l)/2, l=0,1.
\end{equation}
Then we use the motion vectors with corresponding resolutions to perform motion compensation (warp) to each channel of the multi-scale bi-directional reference features. 
\begin{equation}
    \Tilde{F}_f^l= warp(\hat{F}_f^l, \hat{v}_f^l), l=0,1,2.
\end{equation}
\begin{equation}
    \Tilde{F}_b^l= warp(\hat{F}_b^l, \hat{v}_b^l), l=0,1,2.
\end{equation}
Finally, we fuse (FU) the warped multi-scale bi-directional features and obtain multi-scale bi-directional temporal contexts. 
\begin{equation}
    C_f^l= FU_f(\Tilde{F}_f^l), l=0,1,2.
\end{equation}
\begin{equation}
    C_b^l= FU_b(\Tilde{F}_b^l), l=0,1,2.
\end{equation}
When the forward or backward reference frame is an I-frame, as shown in Fig~\ref{fig:reference_propagation}(a)(b)(c), we first use a convolutional layer to transform it into the feature domain and then perform the abovementioned temporal context mining.
The detailed architecture of the temporal context mining module can be found in~\cite{sheng2022temporal}. Different from~\cite{yang2024ucvc,yilmaz2021end,chen2023b}, we do not merge the bi-directional temporal contexts.

\subsection{Bi-directional Contextual Compression}~\label{sec:contextual_ED}
To make full use of temporal predictions, we feed the multi-scale bi-directional temporal contexts ($C_f^0$, $C_f^1$, $C_f^2$), ($C_b^0$, $C_b^1$, $C_b^2$) into the contextual encoder-decoder, as shown in Fig.~\ref{fig:CED}. At the entrance to the contextual encoder, the channel-wise concatenation operation is performed between the bi-directional temporal contexts ($C_f^0$, $C_b^0$) and input frame $x_t$, instead of the traditional subtraction operation. The contextual encoder can learn how to reduce the temporal redundancy automatically. Considering that the temporal redundancy may not be sufficiently reduced at the entrance to the contextual encoder, we further concatenate the bi-directional temporal contexts ($C_f^1$, $C_b^1$), ($C_f^2$, $C_b^2$) with the intermediate features with different resolutions of the contextual encoder. With the spatial non-linear transform consisting of convolutional layers and bottleneck residual blocks~\cite{sheng2022temporal}, the input frame  $x_t$ can be compressed into a compact latent representation $\hat{y}_t$. In the contextual decoder, we also concatenate the bi-directional temporal contexts ($C_f^0$, $C_f^1$, $C_f^2$), ($C_b^0$, $C_b^1$, $C_b^2$) with the intermediate features with different resolutions to complement the temporal information for video reconstruction. To enhance the video reconstruction ability, we follow~\cite{li2023neural, li2022hybrid} and insert two U-shaped blocks into the contextual decoder. Before obtaining the reconstructed frame $\hat{x}_{t}$, we regard a decoded feature $\hat{F}_t$ before the last convolutional layer as the reference feature to help compress the next frame.  Similar to the motion encoder-decoder, to make our model support variable bitrates, we insert two learnable quantization steps ($q_{dec}^{y}$,$q_{enc}^{y}$)~\cite{sheng2024spatial,li2023neural} into the contextual encoder and decoder. 
\begin{figure}[t]
  \centering
   \includegraphics[width=\linewidth]{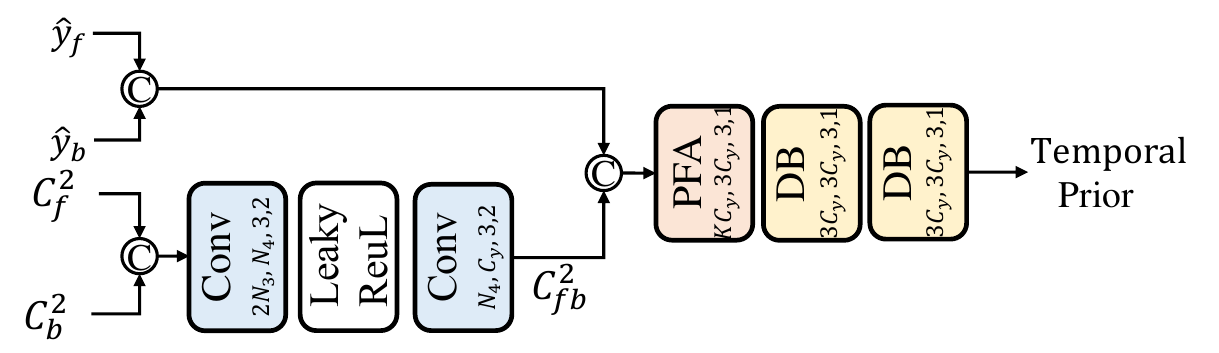}
      \caption{Architecture of the bi-directional temporal entropy model.}
   \label{fig:TE}
\end{figure}
\begin{figure*}[t]
  \centering
   \includegraphics[width=\linewidth]{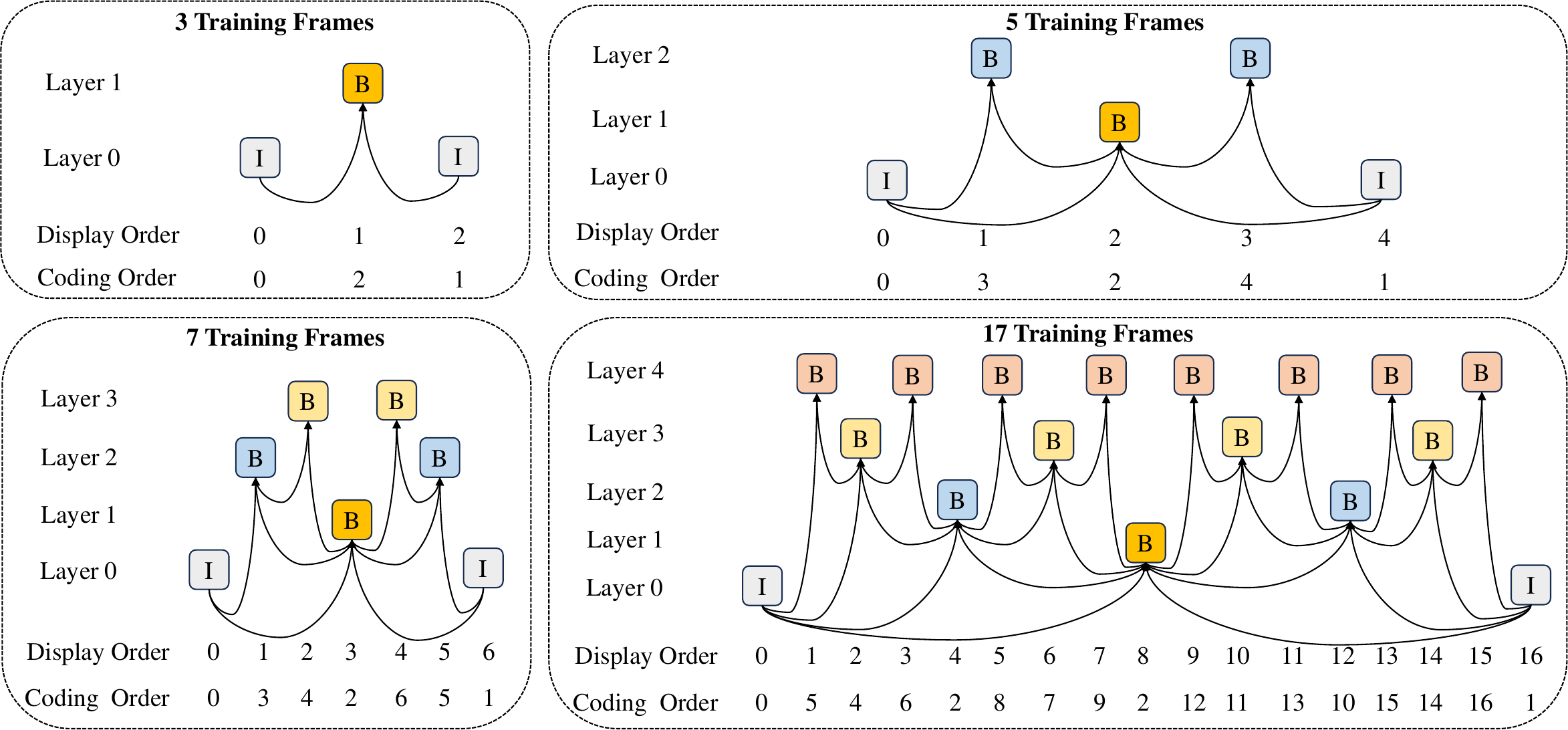}
      \caption{Training reference structures for different numbers of training frames.}
   \label{fig:training_reference}
\end{figure*}
\subsection{Bi-directional Temporal Entropy Model}\label{sec:TEM}
To further utilize the temporal predictions, we propose a bi-directional temporal entropy model, as shown in Fig.~\ref{fig:TE}. The temporal entropy model learns a temporal prior from the smallest bi-directional temporal contexts ($C_f^{2}$, $C_b^{2}$) and the bi-directional decoded latent representations ($\hat{y}_f$, $\hat{y}_b$). We first use two convolutional layers to obtain a fused feature $C_{fb}^2$ with the same resolution and the same channel number as the current latent representation $\hat{y}_t$ from the bi-directional temporal contexts ($C_f^{2}$, $C_b^{2}$). Then we use a prior feature adaptor (PFA) and two depth blocks to fuse $C_{fb}^2$ with the bi-directional decoded latent representations ($\hat{y}_f$, $\hat{y}_b$). The prior feature adaptor is also implemented by a depth block. We use different prior feature adaptors for different types of latent representation propagation. 
When both the forward and backward reference frames are I-frames, as shown in Fig.~\ref{fig:reference_propagation}(a), the $PFA_0$ is used. No decoded latent representation is fused.
\begin{equation}
    PF_{out}= PFA_0(C_{fb}^2).
\end{equation}
When the forward reference frame is I-frames and the backward reference frame is B-frame, as shown in Fig.~\ref{fig:reference_propagation}(b), the $PFA_1$ is used. The latent representation $\hat{y}_b$ of the backward reference frame is fused.
\begin{equation}
    PF_{out}= PFA_1(concat(C_{fb}^2,\hat{y}_b)).
\end{equation}
When the forward reference frame is B-frames and the backward reference frame is I-frame, as shown in Fig.~\ref{fig:reference_propagation}(c), the $PFA_2$ is used. The latent representation $\hat{y}_f$ of the forward reference frame is fused.
\begin{equation}
    PF_{out}= PFA_1(concat(C_{fb}^2,\hat{y}_f)).
\end{equation}
When the forward and backward reference frames are both B-frames, as shown in Fig.~\ref{fig:reference_propagation}(d), the $PFA_3$ is used. The latent representations ($\hat{y}_f$, $\hat{y}_b$)  of the forward and backward reference frames are fused.
\begin{equation}
    PF_{out}= PFA_3(concat(C_{fb}^2,\hat{y}_f, \hat{y}_b)).
\end{equation}
After obtaining the fused temporal prior,  we feed it with the hyperprior into the quadtree partition-based spatial context model~\cite{sheng2024spatial,li2023neural} for entropy modeling.

\begin{table}[t]
\caption{Training strategy of our proposed deep B-frame coding scheme.}
\centering
\begin{tabular}{c|c|c|c|c}
\toprule[1.5pt]
Frames  & Network   & Loss                          &Learning Rate        &Epoch \\ \hline
3     & Inter   & $ L_{t}^{meD}$                    &$1e-4$     & 2      \\ \hline
3     & Recon   & $ L_{t}^{recD}$                   &$1e-4$     & 1      \\ \hline
5     & Recon   & $ L_{t}^{recD}$                   &$1e-4$     & 1      \\ \hline
7     & Recon   & $ L_{t}^{recD}$                   &$1e-4$     & 1      \\ \hline
7     & Inter   & $ L_{t}^{meD}$                    &$1e-4$     & 2      \\ \hline
7     & Inter   & $ L_{t}^{meRD}$                   &$1e-4$     & 6      \\ \hline
7     & Recon   & $ L_{t}^{recD}$                   &$1e-4$     & 2      \\ \hline
7     & Recon   & $ L_{t}^{recRD}$                  &$1e-4$     & 6      \\ \hline
7     & All     & $ L_{t}^{all}$                    &$1e-4$     & 4      \\ \hline
7     & All     & $ L_{t}^{all}$                    &$5e-5$     & 3      \\ \hline
7     & All     & $ L_{t}^{all}$                    &$1e-5$     & 2      \\ \hline
7     & All     & $ L_{t}^{all}$                    &$5e-6$     & 2      \\ \hline
17    & All     & $ L_{t}^{all}$                    &$5e-6$     & 2      \\ 
\bottomrule[1.5pt]
\end{tabular}
\label{table:training_stategy_rgb}
\end{table}
\begin{figure*}[t]
  \centering
   \includegraphics[width=\linewidth]{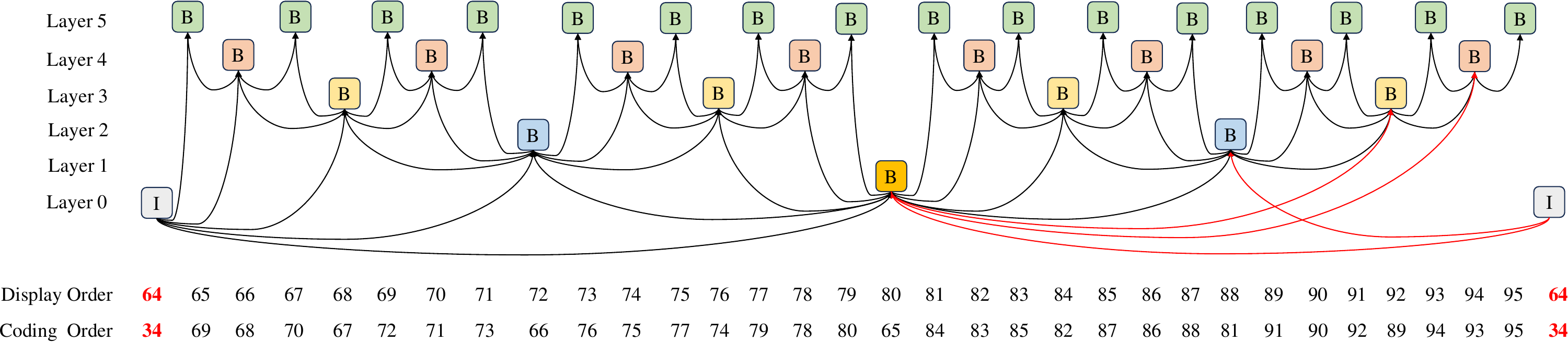}
      \caption{Structure of the last GOP of our proposed DCVC-B scheme when compressing 96 frames. The structure is the same as VTM with the default \emph{encoder\_randomaccess\_vtm} configuration (VTM-RA-GOP32) except that it has only 2 reference frames (VTM-RA-GOP32 has 4 reference frames). }
   \label{fig:last_GOP_structure}
\end{figure*}
\subsection{Hierarchical Quality Structure}~\label{training_strategy}
To achieve a hierarchical quality structure for B-frame coding, traditional video codecs commonly assign different QPs (quantization parameters) to video frames of different temporal layers~\cite{schwarz2006analysis}. Recent deep P-frame coding schemes~\cite{sheng2024spatial,li2023neural} periodically increase the quality of video frames by assigning a quality coefficient that varies periodically over the frame index to the Lagrangian multiplier $\lambda$ in the rate-distortion (R-D) loss function. Inspired by traditional video codecs and deep P-frame coding schemes, we propose to assign different quality coefficients $w_t$ for the B-frames of different temporal layers. As shown in Fig.~\ref{fig:GOP_structure}, we set $w_t$ of the B-frames in Layer 1, 2, 3, 4, and 5 to [1.4, 1.4, 0.7, 0.5, 0.5], respectively. When calculating the loss functions to train our model, we multiply $w_t$ on the $\lambda$. In addition, to make our model adapt to different qualities better, a quality adaptor implemented by one $1\times1$ convolutional layer is inserted before the feature extractor module of the temporal context mining. For B-frames of different temporal layers, different quality adaptors are selected. Note that the quality adaptors are included in the training from the beginning. \par
We combine the hierarchical quality structure-based training strategy with the step-by-step training strategy. The detailed training strategies are listed in Table.~\ref{table:training_stategy_rgb}. For different numbers of training frames, the training reference structures are illustrated in Fig.~\ref{fig:training_reference}. The loss functions for different steps are defined as follows.

\begin{itemize}
\item $L_{t}^{meD}$ refers to the prediction distortion $D_{t}^m$ between $x_t$ and its predicted frame $\tilde{x}_t$. We use the decoded bi-directional motion vectors ($\hat{v}_{t\rightarrow f}$, $\hat{v}_{t\rightarrow b}$) to perform motion compensation to the bi-directional reference frames ($\hat{x}_{f}$, $\hat{x}_{b}$). Then we use a mask network~\cite{yilmaz2021end} implemented by two depth blocks to merge the bi-directional predicted frames ($\tilde{x}_{f}$, $\tilde{x}_{b}$), generating the final predicted frame $\tilde{x}_t$. Note that, the mask network is only used in the training process.
\begin{equation}
    L_{t}^{meD}= w_t \cdot \lambda \cdot D_{t}^m.
\label{loss1}
\end{equation}
\item $L_{t}^{meRD}$ refers to the trade-off between the prediction distortion $D_{t}^m$ and motion coding bitrate $R_{t}^{m}$. 
\begin{equation}
    L_{t}^{meRD}= w_t \cdot \lambda \cdot D_{t}^{m} + R_{t}^{m}.
\label{loss2}
\end{equation}

\item $L_{recD}$ refers to the reconstruction distortion $D_{t}^{y}$ between $x_t$ and its decoded frame $\hat{x}_t$.
\begin{equation}
    L_{t}^{recD}= w_t \cdot \lambda \cdot D_{t}^{y}.
\label{loss3}
\end{equation}

\item $L_{t}^{recRD}$ refers to the trade-off between the reconstruction distortion $D_{t}^{y}$ and the contextual coding bitrate $R_{t}^{y}$.
\begin{equation}
    L_{t}^{recRD}= w_t \cdot \lambda \cdot D_{t}^{y} + R_{t}^{y}.
\label{loss4}
\end{equation}

\item $L_{t}^{all}$ refers to the trade-off between the reconstruction distortion $D_{t}^{y}$ and all the consumed bitrate.
\begin{equation}
\begin{aligned}
    L_{t}^{all}&= w_t \cdot \lambda \cdot D_{t}^{y} + R_{t}^{m} + R_{t}^{y}.
\end{aligned}
\label{loss5}
\end{equation}
\end{itemize}
Unlike existing deep P-frame coding schemes~\cite{sheng2022temporal,li2022hybrid,li2023neural,li2024neural,sheng2024spatial}, we do not calculate the average loss for multi-frame joint training. When using $L_{t}^{meD}$ or $L_{t}^{meRD}$ as the loss function, we only train the motion-related modules (Inter). When using $L_{t}^{recD}$ or $L_{t}^{recRD}$ as the loss function, we only train the context-related modules (Recon). When using $L_{t}^{all}$ as the loss function, we jointly train all the modules (All). 
\par

\begin{figure*}[t]
  \centering
  \begin{minipage}[c]{0.32\linewidth}
  \centering
  \includegraphics[width=\linewidth]{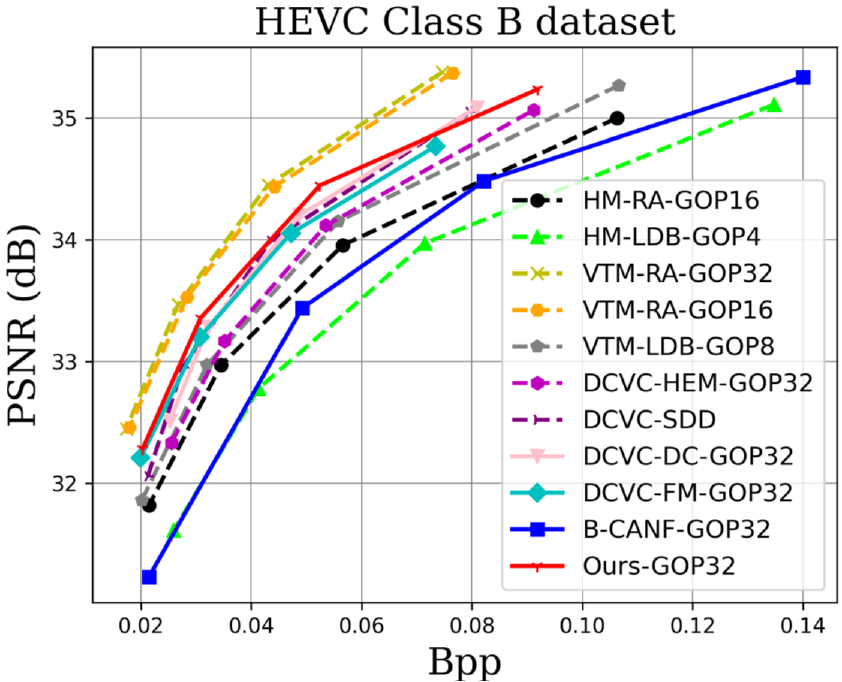}
 \end{minipage}%
  \begin{minipage}[c]{0.32\linewidth}
  \centering
    \includegraphics[width=\linewidth]{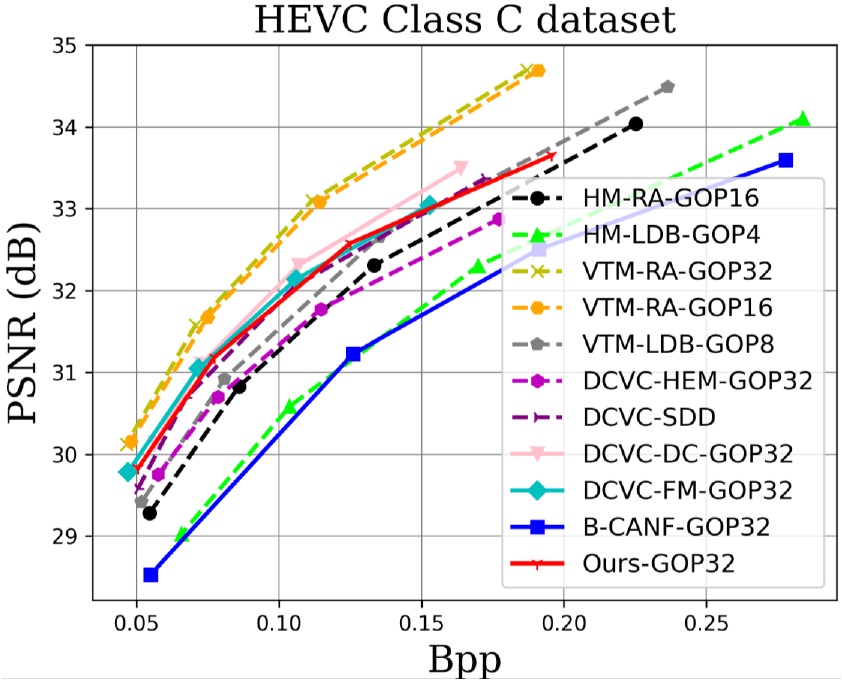}
  \end{minipage}%
  \begin{minipage}[c]{0.32\linewidth}
  \centering
    \includegraphics[width=\linewidth]{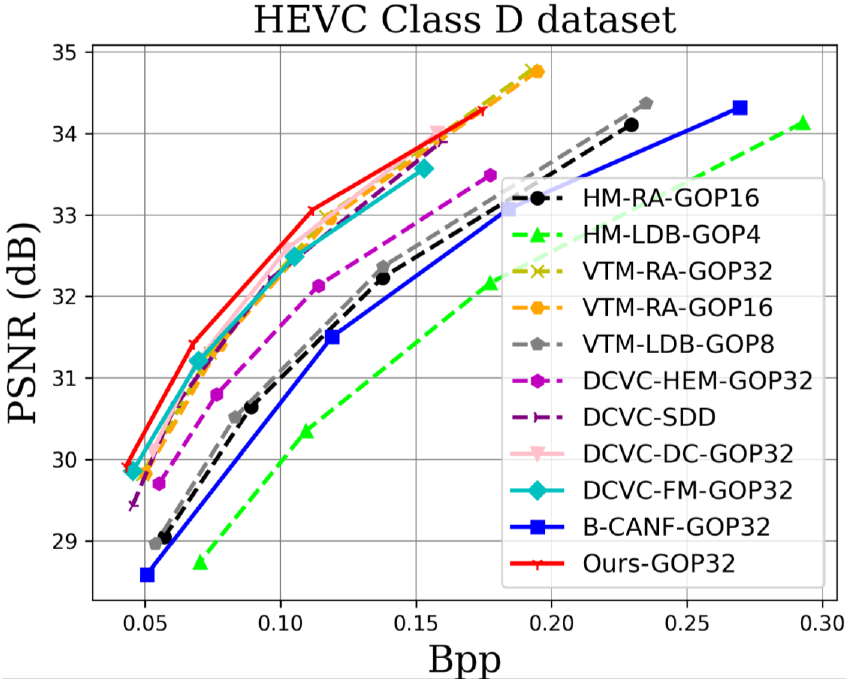}
  \end{minipage}%
  
  \begin{minipage}[c]{0.32\linewidth}
  \centering
    \includegraphics[width=\linewidth]{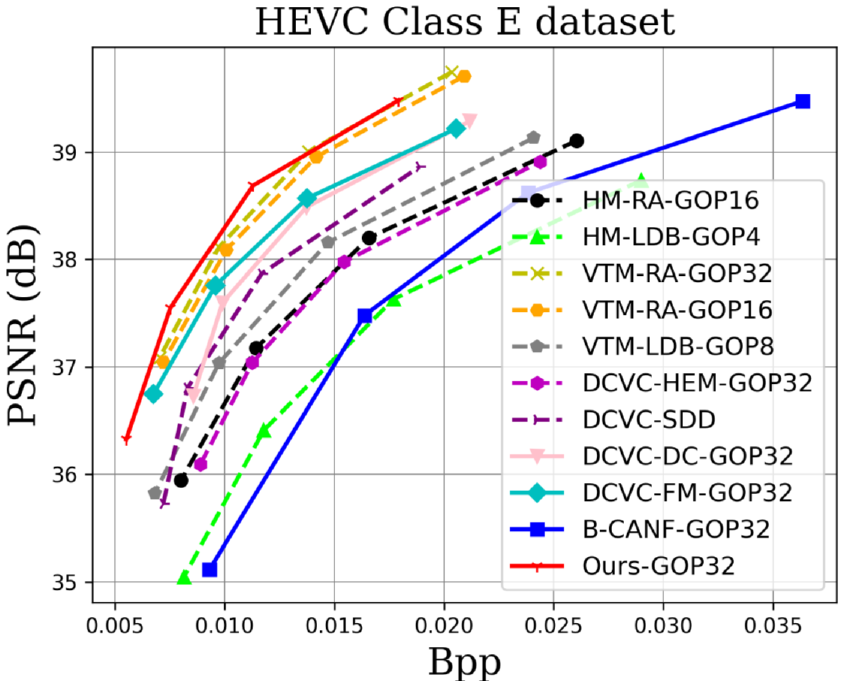}
  \end{minipage}%
  \begin{minipage}[c]{0.32\linewidth}
  \centering
    \includegraphics[width=\linewidth]{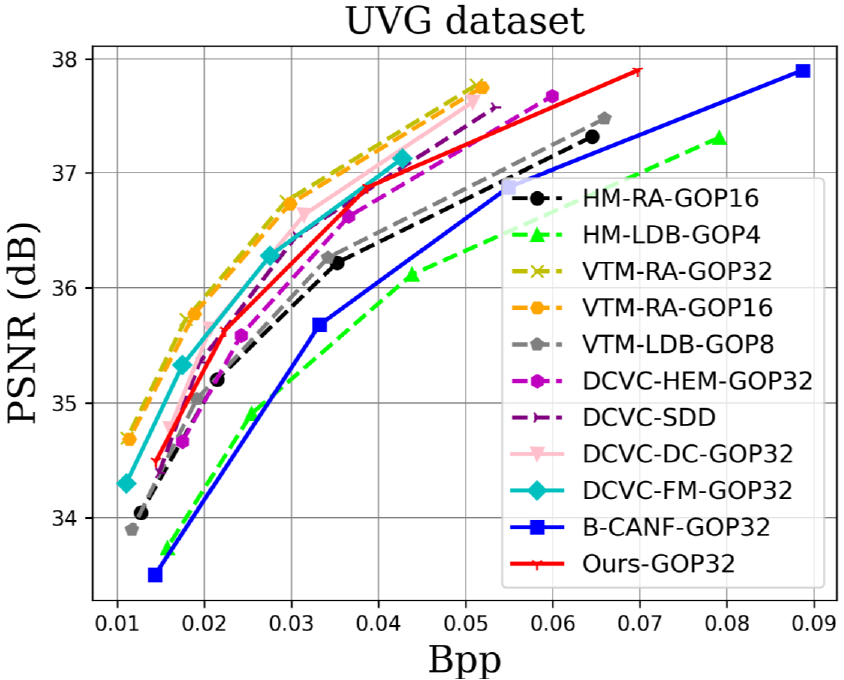}
  \end{minipage}%
  \begin{minipage}[c]{0.32\linewidth}
  \centering
    \includegraphics[width=\linewidth]{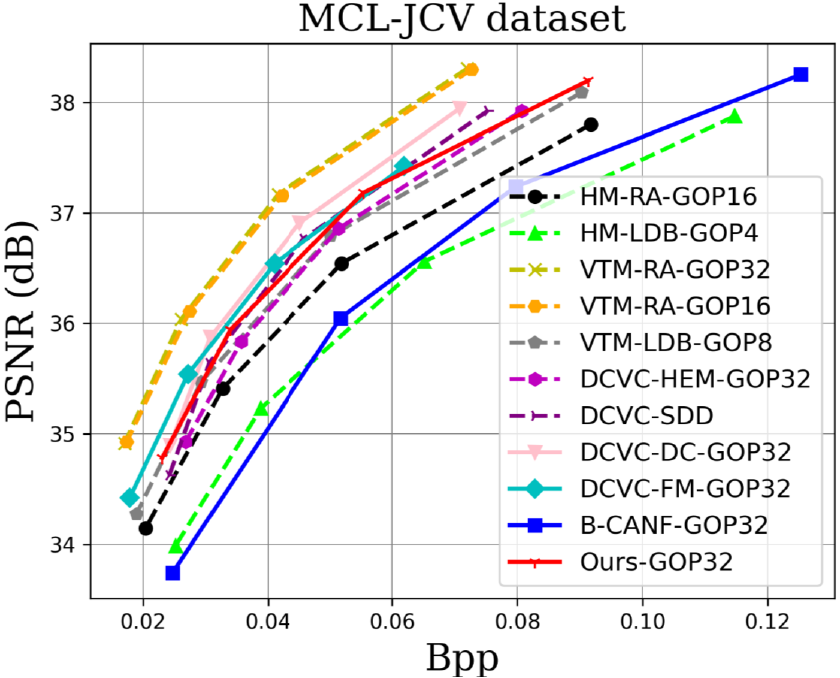}
  \end{minipage}%
    \caption{Rate-distortion curves of the HEVC, UVG, and MCL-JCV video datasets. The reconstruction quality is measured by PSNR.}
  \label{fig:psnr_results}
\end{figure*}
\begin{table*}[t]
\caption{BD-rate (\%) comparison for PSNR. The anchor is HM under the default random access configuration. The intra period is set to 32 for all schemes.} 
  \centering
\scalebox{0.9}{
\begin{threeparttable}
\begin{tabular}{l|c|c|c|c|c|c|c|c}
\toprule[1.5pt]
                        & HEVC Class B  & HEVC Class C  &HEVC Class D &HEVC Class E &HEVC Class RGB  &UVG             &MCL-JCV  & Average\\ \hline
HM-RA-GOP16             &0.0            &0.0            &0.0          &0.0          &0.0             &0.0             &0.0      &0.0     \\ \hline
HM-LDB-GOP4             &23.7           &26.2           &29.7         &32.2         &38.9            &29.8            &23.9     &29.2    \\ \hline
VTM-RA-GOP32            &\bf{--39.1}    &\bf{--33.7}    &--30.4       &--37.9       &\bf{--41.2}          &\bf{--35.9}     &\bf{--38.0}   &\bf{--37.6}   \\ \hline
VTM-RA-GOP16            &--37.2         &--32.1         &--29.2       &--35.4       &--38.8     &--34.0          &--36.9   &--34.8   \\ \hline
VTM-LDB-GOP8            &--11.7         &--8.6          &--3.9        &--10.4       &--3.8           &--5.3           &--13.3   &--8.1    \\ \hline
DCVC-HEM-GOP32          &--10.8         &--0.9          &--15.7       &3.3          &--7.1           &--11.7.        &--10.8   &--7.7    \\ \hline
DCVC-SDD-GOP32          &--24.8         &--10.6         &--27.8       &--17.5       &--20.7          &--24.9         &--19.9   &--20.9    \\ \hline
DCVC-DC-GOP32           &--22.0         &--19.9         &--32.3       &--24.6       &--23.9          &--25.0          &--22.8   &--24.4    \\ \hline
DCVC-FM-GOP32           &--20.5         &--19.3         &--31.1       &--30.5       &--18.1          &--23.7          &--20.7   &--23.4    \\ \hline
B-CANF-GOP32            &13.9           &30.9           &5.1          &28.5         &5.0             &21.0            &22.6     &18.1    \\ \hline
Ours-GOP32              &--26.8         &--15.8         &\bf{--36.7}  &\bf{--42.9}  &--29.3          &--18.0          &--16.8   &--26.6   \\ 
\bottomrule[1.5pt]
\end{tabular}
\end{threeparttable}
}
\label{table:ip32_psnr}
\end{table*}

\begin{figure*}[t]
  \centering
  \begin{minipage}[c]{0.32\linewidth}
  \centering
  \includegraphics[width=\linewidth]{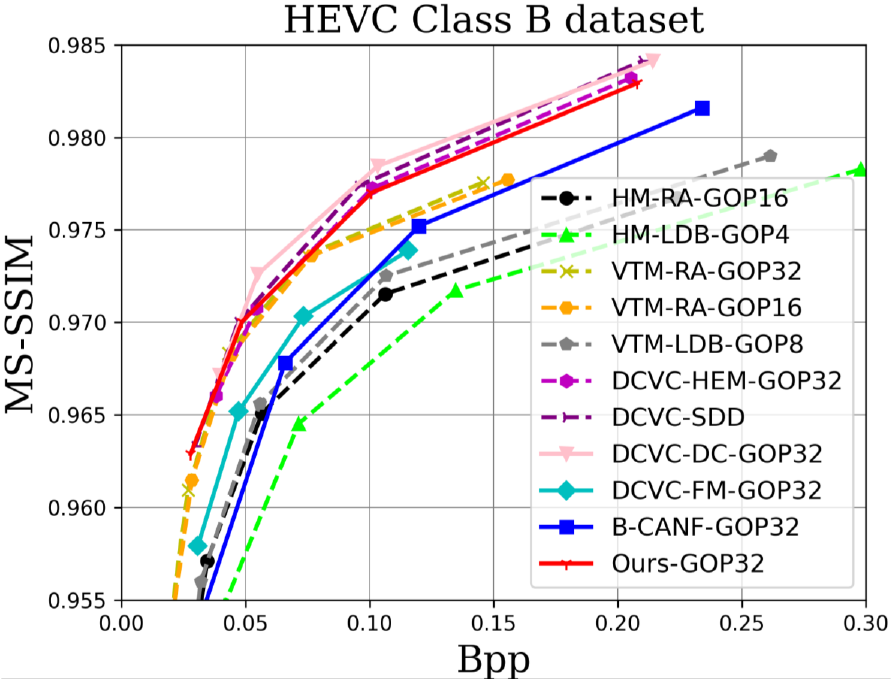}
 \end{minipage}%
  \begin{minipage}[c]{0.32\linewidth}
  \centering
    \includegraphics[width=\linewidth]{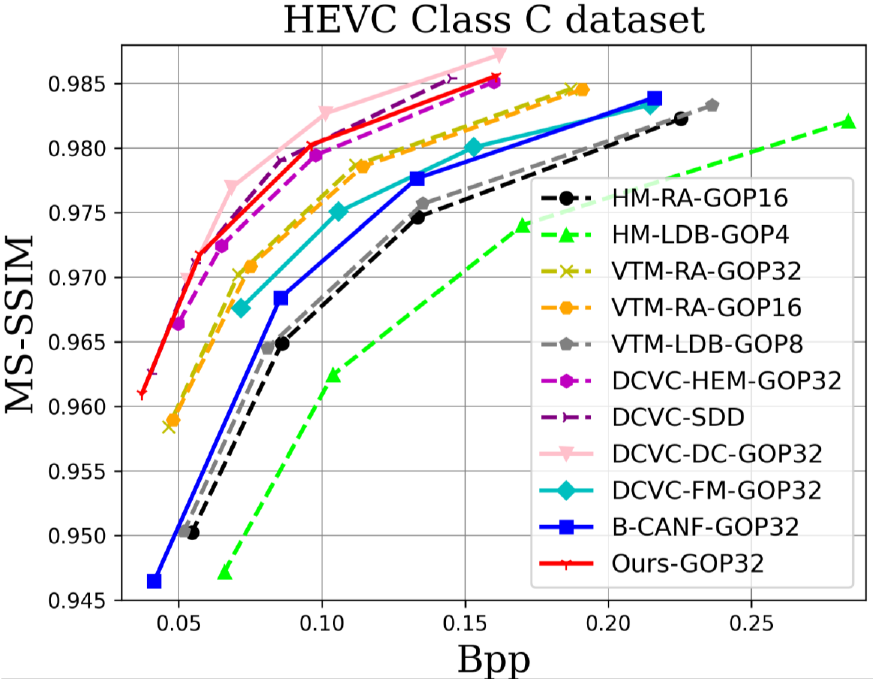}
  \end{minipage}%
  \begin{minipage}[c]{0.32\linewidth}
  \centering
    \includegraphics[width=\linewidth]{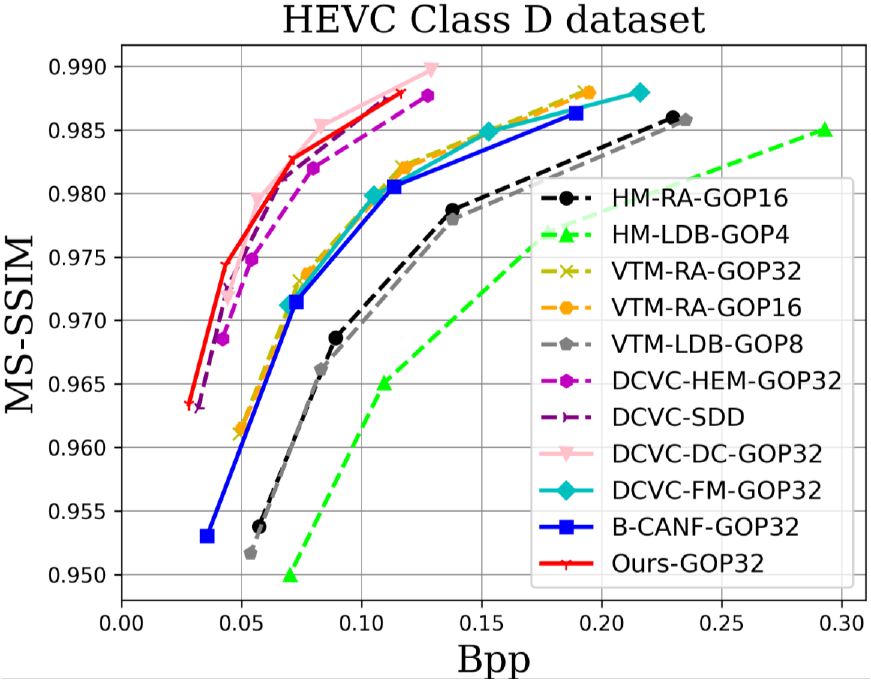}
  \end{minipage}%
  
  \begin{minipage}[c]{0.32\linewidth}
  \centering
    \includegraphics[width=\linewidth]{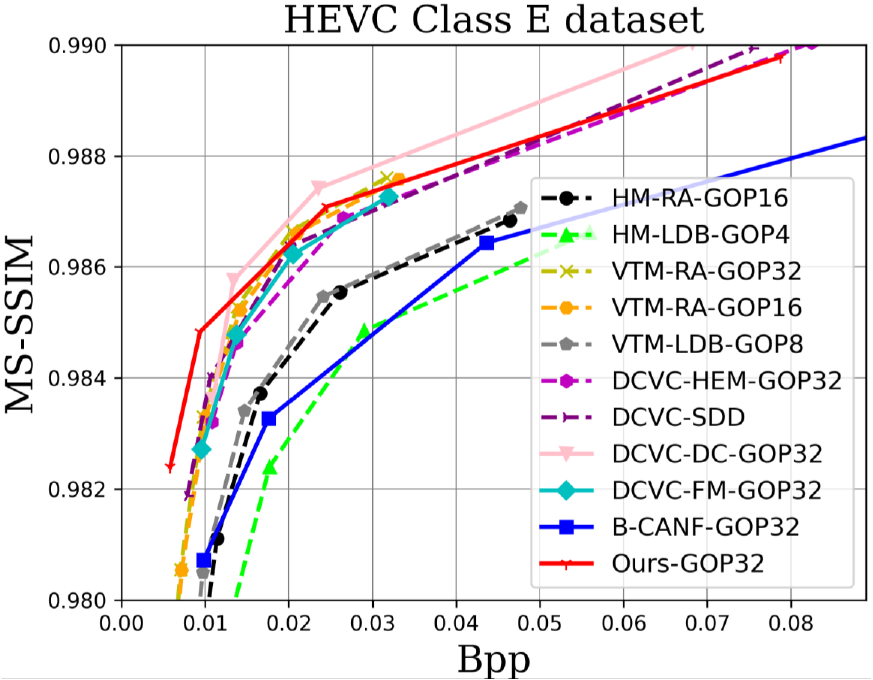}
  \end{minipage}%
  \begin{minipage}[c]{0.32\linewidth}
  \centering
    \includegraphics[width=\linewidth]{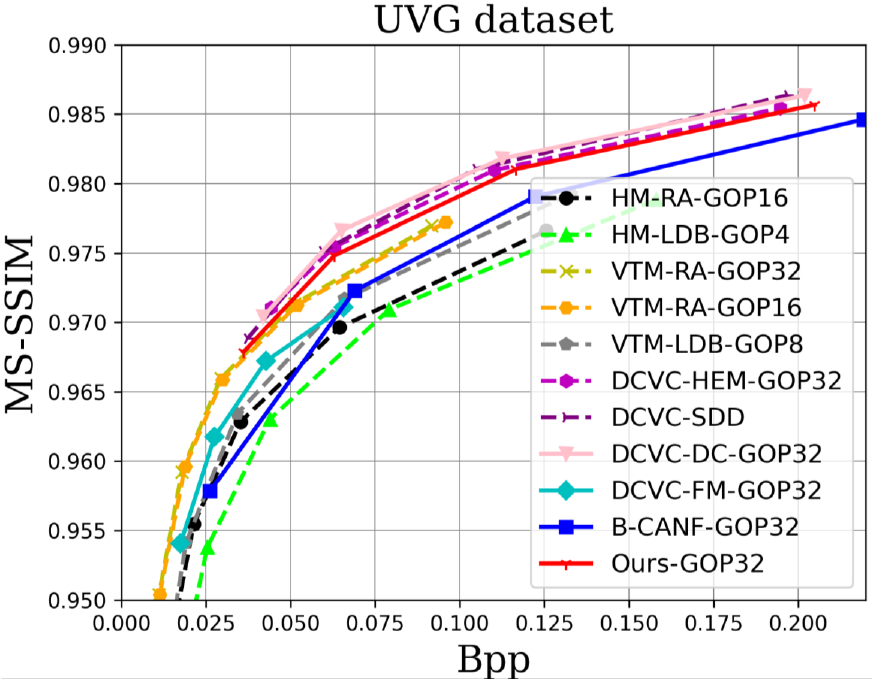}
  \end{minipage}%
  \begin{minipage}[c]{0.32\linewidth}
  \centering
    \includegraphics[width=\linewidth]{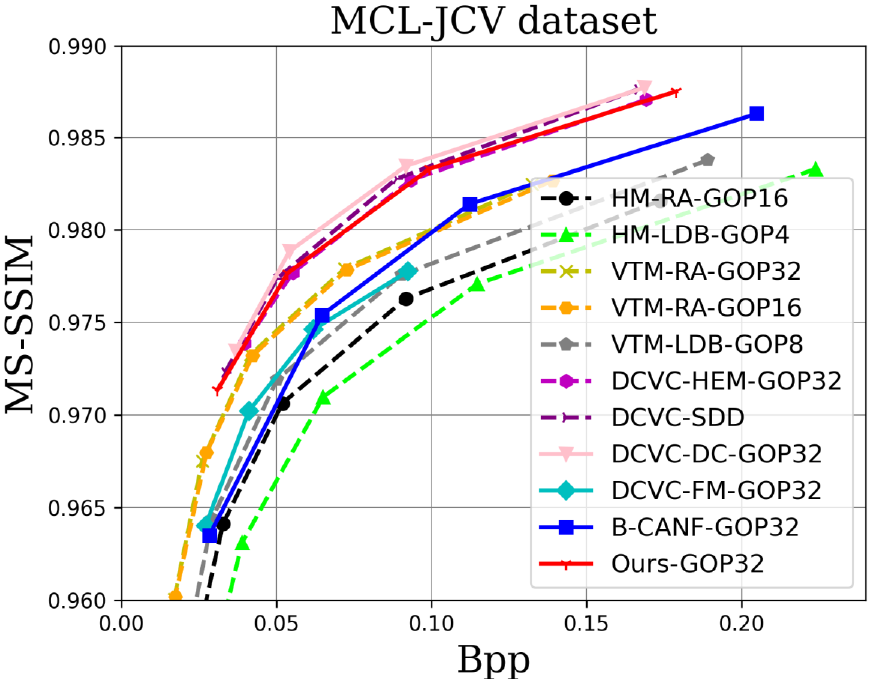}
  \end{minipage}%
    \caption{Rate-distortion curves of the HEVC, UVG, and MCL-JCV video datasets. The reconstruction quality is measured by MS-SSIM.}
  \label{fig:msssim_results}
\end{figure*}
\begin{table*}[t]
\caption{BD-rate (\%) comparison for MS-SSIM. The anchor is HM under the default random access configuration. The intra period is set to 32 for all schemes.} 
  \centering
\scalebox{0.9}{
\begin{threeparttable}
\begin{tabular}{l|c|c|c|c|c|c|c|c}
\toprule[1.5pt]
                        & HEVC Class B  & HEVC Class C  &HEVC Class D &HEVC Class E &HEVC Class RGB  &UVG             &MCL-JCV  & Average\\ \hline
HM-RA-GOP16             &0.0            &0.0            &0.0          &0.0          &0.0             &0.0             &0.0      &0.0     \\ \hline
HM-LDB-GOP4             &29.7           &30.2           &38.1         &30.2         &29.8            &21.8            &21.5     &28.8    \\ \hline
VTM-RA-GOP32            &--35.9         &--32.9         &--30.3       &--34.2       &--38.1          &--33.0          &--36.0   &--34.3   \\ \hline
VTM-RA-GOP16            &--34.3         &--31.7         &--29.6       &--32.5       &--34.9          &--31.2          &--34.9   &--32.7   \\ \hline
VTM-LDB-GOP8            &--5.7          &--4.8          &2.3          &--8.0        &--7.8           &--8.1           &--13.3   &--6.5    \\ \hline
DCVC-HEM-GOP32          &--46.2         &--45.6         &--52.6       &--33.5       &--45.8          &--42.6          &--48.6   &--45.0    \\ \hline
DCVC-SDD-GOP32          &--52.8         &--51.8         &--58.3       &--42.4       &--52.4          &\bf{--45.8}     &\bf{--54.2}   &--51.1    \\ \hline
DCVC-DC-GOP32           &\bf{--53.2}    &\bf{--55.0}    &--60.4       &--48.0       &--\bf{56.6}     &--44.9          &--53.7   &\bf{--53.1}    \\ \hline
DCVC-FM-GOP32           &--17.9         &--21.6         &--27.9       &--35.6       &--23.5          &--14.6          &--17.8   &--22.7    \\ \hline
B-CANF-GOP32            &--9.7          &--15.4         &--28.8       &13.0         &--17.1          &--8.2           &--19.2   &--12.2    \\ \hline
Ours-GOP32              &--47.0         &--50.1         &\bf{--60.9}  &\bf{--55.2}  &--48.9          &--38.4          &--48.9   &--49.9   \\ 
\bottomrule[1.5pt]
\end{tabular}
  \begin{tablenotes}
   \item \footnotesize \dag Due to the MS-SSIM model of DCVC-FM is not released, when the reconstruction quality is measured by MS-SSIM, we use the PSNR model of DCVC-FM to supplement the result, although this is unfair.
  \end{tablenotes}
\end{threeparttable}}
\label{table:ip32_ssim}
\end{table*}

\section{Experiments}\label{sec:experiments} 
\subsection{Experimental Setup}
\subsubsection{Training Dataset}
As listed in Table~\ref{table:training_stategy_rgb}, when the training frame is 3, 5, and 7, we train our model using 7-frame videos of the Vimeo-90k~\cite{xue2019video} dataset for short-sequence training.  When the training frame is 17, we use 9000 33-frame video clips collected from raw Vimeo videos. During training, the video frames are randomly cropped into 256$\times$256 patches.

\subsubsection{Testing Dataset}
To evaluate the performance of our DCVC-B scheme, we use the video sequences from HEVC dataset~\cite{bossen2013common}, UVG dataset~\cite{mercat2020uvg}, and MCL-JCV dataset~\cite{wang2016mcl}. The HEVC datasets contain 22 videos in Class B, C, D, E, and RGB~\cite{sheng2022temporal} with different resolutions from 240p to 1080p. The UVG and  MCL-JCV datasets contain 7 and 30 videos with the resolution of 1080p, respectively.

\subsubsection{Implementation Details}
The DCVC-B scheme allows one model to support variable bitrates by introducing learnable quantization steps, as mentioned in Section~\ref{sec:MCP} and~\ref{sec:contextual_ED}. During training, when the reconstruction quality is measured by PSNR, we set 4 base $\lambda$ values (85, 170, 380, 840) to control the rate-distortion trade-off and set the distortion metrics ($D_{t}^{y}$, $D_t^{m}$) to Mean Squared Error (MSE).
When the reconstruction quality is measured by MS-SSIM, we set the base $\lambda$ values to ($\frac{85}{17}$, $\frac{170}{17}$, $\frac{380}{17}$, $\frac{840}{17}$) and set the distortion metrics ($D_{t}^{y}$, $D_t^{m}$) to 1--MS-SSIM. Since existing deep video coding schemes have outperformed traditional video codecs by a large margin regarding MS-SSIM, we only fine-tune our MS-SSIM model for 2 epochs based on the PSNR model. During testing, we can interpolate the quantization steps to achieve other bitrates. PyTorch is used to implement our model. AdamW~\cite{kingma2014adam} is used as the optimizer and the batch size is set to 8. 

\begin{figure*}[t]
  \centering
   \includegraphics[width=\linewidth]{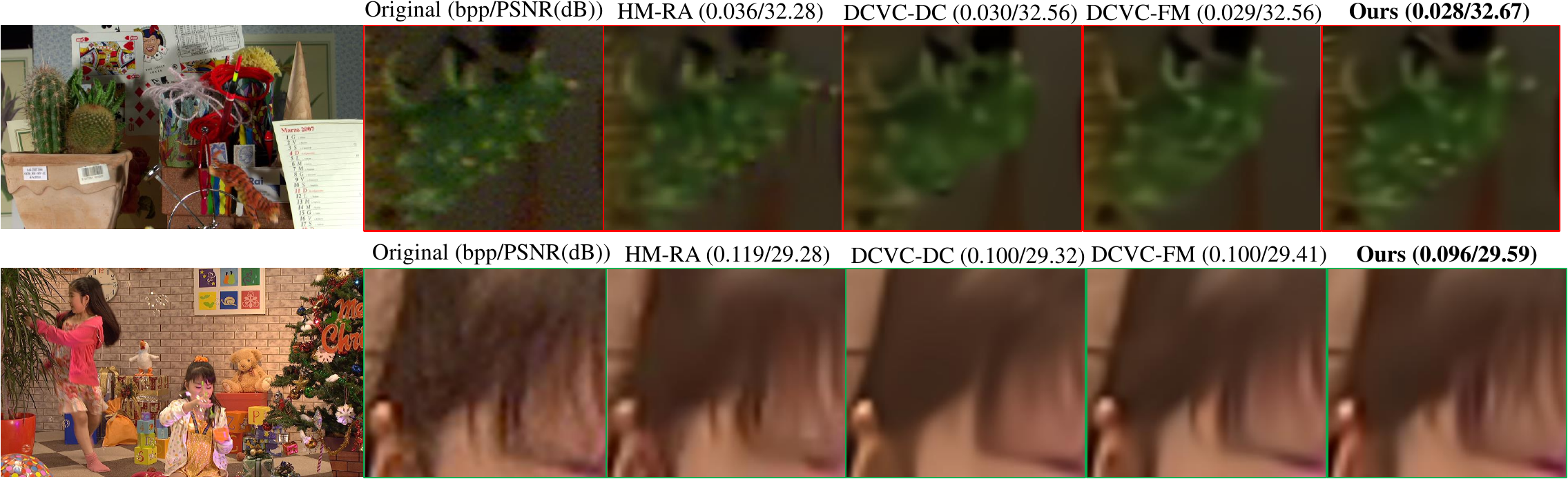}
      \caption{Subjective quality comparison on the \emph{Cactus\_1920x1080\_50} sequence in the HEVC Class B dataset and the \emph{PartyScene\_832x480\_50} sequence in the HEVC Class C dataset.}
   \label{fig:subjective}
\end{figure*}
\subsubsection{Test Configurations}
In this paper, we focus on the random access (RA) scenario.
For traditional video codecs, we choose the reference software of H.265/HEVC---HM-16.20~\cite{HM} and the reference software of H.266/VVC---VTM-13.2~\cite{VTM} as our benchmarks. For HM, we use the default \emph{encoder\_randomaccess\_main\_rext} configuration (Intra period 32, GOP size 16). We denote HM under this configuration as HM-RA-GOP16. For VTM, we use the default \emph{encoder\_randomaccess\_vtm} (Intra period 32, GOP size 32) and \emph{encoder\_randomaccess\_vtm\_gop16.cfg} (Intra period 32, GOP size 16) configurations.  We denote VTM under these configurations as VTM-RA-GOP16 and VTM-RA-GOP32. \par
To show the compression performance gap between the traditional video codecs under random access configuration and low delay (LD) configuration, we also test HM-16.20 and VTM-13.2 under \emph{encoder\_lowdelay\_main\_rext} and \emph{encoder\_lowdelay\_vtm} configurations. The GOP size of HM and VTM are set to 4 and 8 by default. Their intra periods are set to 32.  We denote HM and VTM under these configurations as HM-LDB-GOP4 and VTM-RA-GOP8. \par
\par
For neural video coding models, we compare with existing state-of-the-art deep P-frame coding schemes, including DCVC-HEM~\cite{li2022hybrid}, DCVC-SDD~\cite{sheng2024spatial}, DCVC-DC~\cite{li2023neural}, and DCVC-FM~\cite{li2024neural}. We also compare with existing deep B-frame coding scheme B-CANF~\cite{chen2023b}. For our scheme, we follow the existing deep P-frame coding schemes~\cite{li2022hybrid, sheng2024spatial,li2023neural,li2024neural} and compress 96 frames of one video to make a fair comparison. For B-CANF, we use its default setting and compress 97 frames but calculate BD-rate using the compression results of the first 96 frames. The intra period and GOP size of our DCVC-B scheme and existing deep coding schemes are all set to 32. When compressing 96 frames, the structure of the first two GOPs of our codec is shown in Fig.~\ref{fig:GOP_structure} and the structure of the last GOP is shown in Fig.~\ref{fig:last_GOP_structure}, which is the same as the VTM-RA-GOP32 except that it has only 2 reference frames (VTM-RA-GOP32 has 4 reference frames). We use the same I-frame codec as DCVC-DC.

\subsection{Experimental Results}
\subsubsection{Comparison Results}
We illustrate the RD-curves in terms of PSNR and MS-SSIM for different testing datasets in Fig.~\ref{fig:psnr_results} and Fig.~\ref{fig:msssim_results}. We also report the corresponding BD-rate values in 
Table~\ref{table:ip32_psnr} and Table~\ref{table:ip32_ssim}. The anchor is HM-RA-GOP16. Negative values indicate bitrate saving compared with HM while positive values indicate bitrate increasing. The comparison results show that in terms of PSNR, our proposed DCVC-B scheme obtains an average --26.6\% BD-rate reduction against the anchor HM-RA-GOP16, which even outperforms the SOTA deep P-frame coding schemes DCVC-DC (--24.4\%) and DCVC-FM (--23.4\%). For the HEVC Class D and Class E datasets, we even outperform VTM-RA-GOP32. Compared with the state-of-the-art deep B-frame coding scheme B-CANF~\cite{chen2023b}, our DCVC-B scheme achieves a large compression performance improvement. In terms of MS-SSIM, DCVC-B obtains an average --49.9\% BD-rate reduction against HM-RA-GOP16 and even outperforms VTM-RA-GOP32 setting (--34.3\%) over all the datasets.  The subjective comparison results illustrated in Fig.~\ref{fig:subjective} show that the reconstructed videos of our scheme can retain more details.\par
\begin{table}[t]
 \centering
 \caption{Runtime and computational complexity comparison for a 1080p video frame.}
\scalebox{1}{
\begin{tabular}{c|c|c|c|c}
\toprule[1.5pt]
Schemes  & Enc Time & Dec Time & MACs/pixel & Model Size           \\ \hline
HM-RA  & 92.24 s& 0.29 s &---&---\\ \hline
VTM-RA  & 1144.70 s& 0.37 s &----&---\\ \hline
DCVC-HEM & 0.75 s& 0.26 s & 1791.64K& 17.52M\\ \hline
DCVC-SDD & 0.94 s& 0.74 s & 1849.06K& 18.74M\\ \hline
DCVC-DC     & 0.82 s& 0.64 s & 1397.90K&18.45M\\ \hline
DCVC-FM& 0.74 s& 0.53 s &1180.77K &17.02M\\ \hline
B-CANF&  1.49 s& 1.06 s &3081.11K &23.66M\\ \hline
Ours     & 1.19 s& 0.99 s  &3004.52K &22.28M\\
\bottomrule[1.5pt]
\end{tabular}}
\label{time}
\end{table}
\subsubsection{Runtime and Computational Complexity Comparison}
We list the encoding and decoding time for 1920$\times$1080 videos of different schemes in Table~\ref{time}. When calculating the encoding and decoding time of deep video compression schemes, we follow the setting of~\cite{sheng2022temporal,sheng2024spatial,li2023neural,li2022hybrid} and include the time for model inference, entropy modeling, entropy coding, and data transfer between CPU and GPU. We also compare the computational complexities and model sizes of deep video compression schemes in Table~\ref{time}. We run deep video compression schemes on an NVIDIA 3090 GPU and run traditional video codecs on an Intel(R) Xeon(R) Gold 5118 CPU. The comparison results show that the complexity of our proposed DCVC-B scheme is increased compared with existing deep P-frame coding schemes. The main reason is the increased complexity of bi-directional motion estimation on both the encoder and decoder. We will try our best to reduce the complexity in the future. \par

\subsection{Ablation Studies}
\subsubsection{Effectiveness of Proposed Technologies}
We conduct an ablation study to verify the effectiveness of our proposed technologies on the HEVC dataset. We progressively add our proposed technologies to the baseline $M_0$ and design four models ($M_1$, $M_2$, $M_3$, $M_4$), as listed in Table~\ref{effectiveness}. By comparing $M_0$ and $M_1$, we find that the bi-directional motion difference context propagation (BMDCP) can bring 5.0\% BD-Rate reduction, which shows that the BMDCP can efficiently reduce the motion coding costs. Based on $M_1$, the model $M_2$ and $M_3$ with the bi-directional contextual compression model (BCCM) and the bi-directional temporal entropy model (BTEM) can improve the performance gain to 43.9\%, which indicates that the BCCM and BTEM can make better use of the temporal correlation. Based on $M_3$, our complete model $M_4$ with the hierarchical quality structure-based training strategy can further bring an additional 10.9\% bitrate saving, which verifies the effectiveness of our proposed training strategy. 
 \begin{table}[t]
\caption{Effectiveness of proposed technologies.}
\centering
\scalebox{1}{
\begin{tabular}{c|c|c|c|c|c}
\toprule[1.5pt]
Model Index         &BMDCP        & BCCM        &BTEM &HQS & BD-Rate (\%)\\ \hline
$M_0$&\XSolidBrush  &\XSolidBrush &\XSolidBrush &\XSolidBrush & 0.0  \\ \hline
$M_1$&\Checkmark    &\XSolidBrush &\XSolidBrush &\XSolidBrush &--5.0 \\ \hline
$M_2$&\Checkmark    &\Checkmark   &\XSolidBrush &\XSolidBrush & --37.2    \\\hline
$M_3$&\Checkmark    &\Checkmark   &\Checkmark   &\XSolidBrush &--43.9    \\\hline
$M_4$&\Checkmark    &\Checkmark   &\Checkmark   &\Checkmark   & --54.8    \\
\bottomrule[1.5pt]
\end{tabular}
}
\label{effectiveness}
\end{table}

\subsubsection{Effectiveness of Bi-directional Motion Difference Context Propagation}
\begin{figure}[t]
  \centering
   \includegraphics[width=0.8\linewidth]{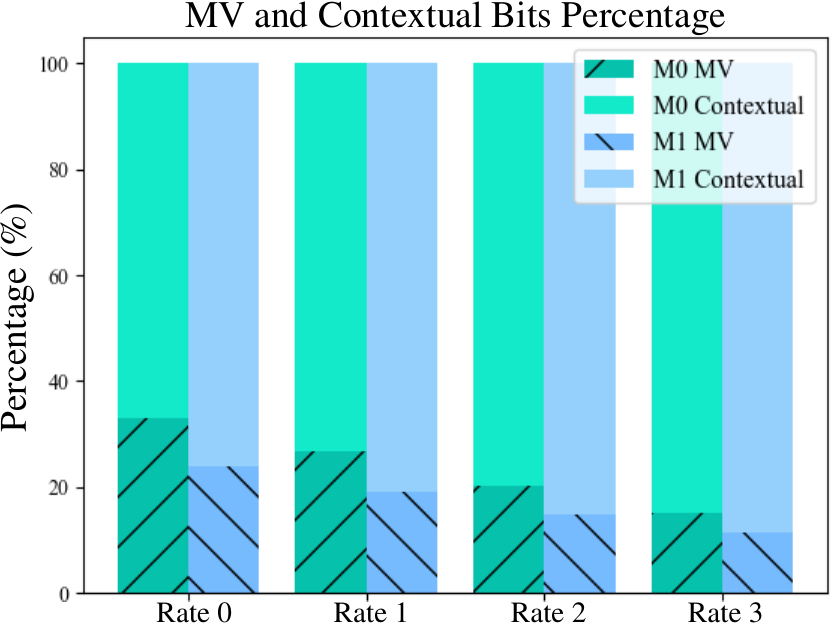}
      \caption{Percentages of MV bits and contextual bits comparison of different models on the HEVC dataset. Rate 0 is the lowest bitrate point and Rate 3 is the highest bitrate point.}
   \label{ablation:bits}
\end{figure}

To analyze why our proposed bi-directional motion difference context propagation method can bring performance gain, we compare the MV and contextual bit percentages of $M_0$ and $M_1$ models. As illustrated in Fig.~\ref{ablation:bits}, we find that the proposed BMDCP method can efficiently reduce the MV bits percentages for different bitrates. For example, for the lowest bitrate (Rate 0), the MV of $M_0$ model accounts for 33.1\% of the total bitrate, while the MV of $M_1$ model only accounts for 23.8\% of the total bitrate.
\begin{figure*}[t]
  \centering
   \includegraphics[width=\linewidth]{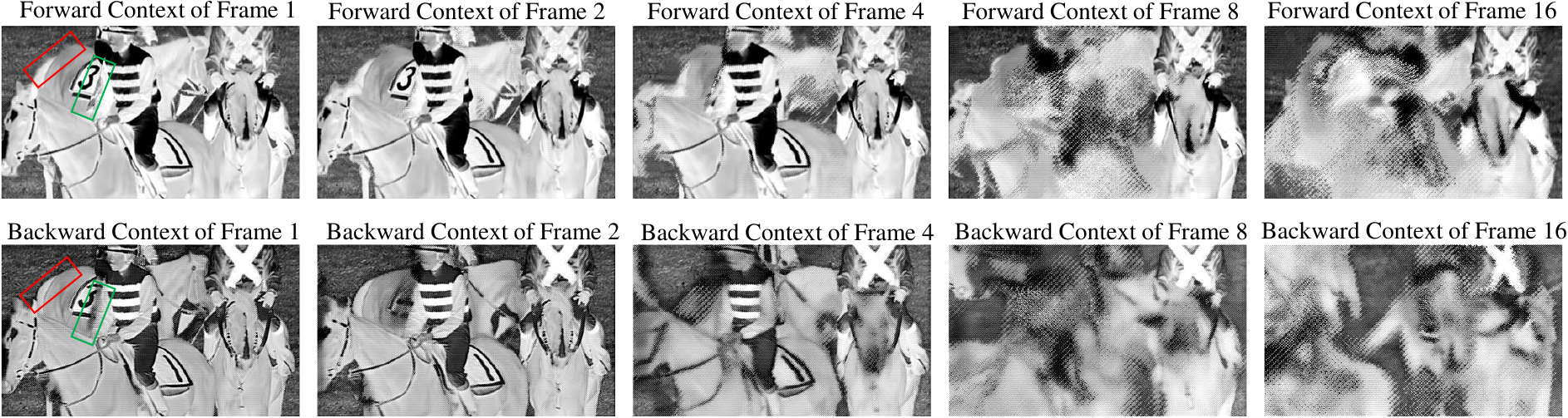}
      \caption{Visualization of the forward and backward temporal contexts of different B-frames.}
   \label{ablation:tcm}
\end{figure*}
\begin{figure}[t]
  \centering
   \includegraphics[width=\linewidth]{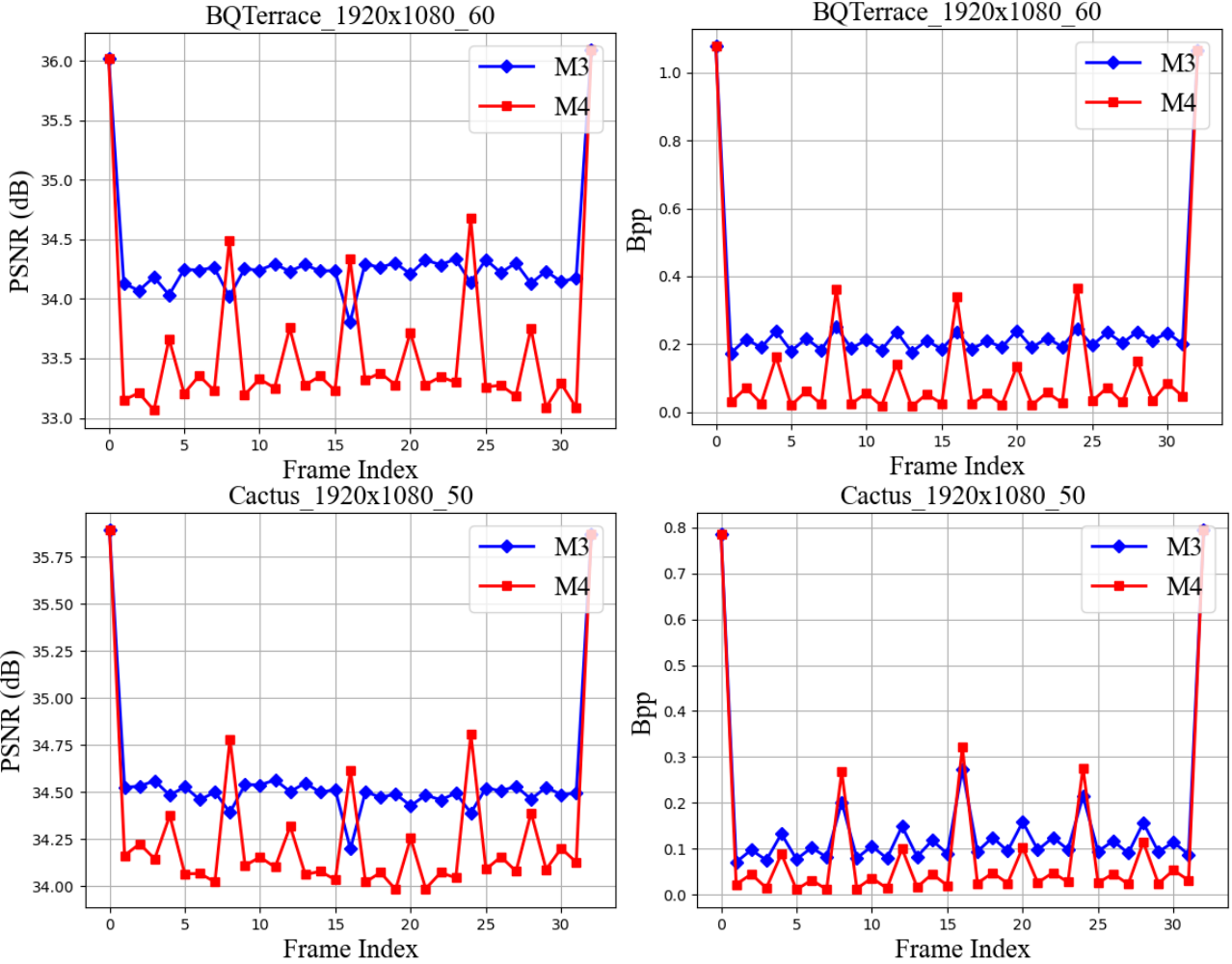}
      \caption{Frame quality and bitrate comparison of different models on the HEVC Class B dataset.}
   \label{ablation:hqs}
\end{figure}
\begin{table}[t]
 \centering
 \caption{Influence of the values of hierarchical quality coefficients.}
\scalebox{1}{
\begin{tabular}{c|c}
\toprule[1.5pt]
Quality Coefficients  & BD-Rate (\%)\\ \hline
[1.0, 1.0, 1.0, 1.0, 1.0] &0.0\\ \hline
[1.4, 1.4, 1.4, 1.4, 1.4] &--4.7\\ \hline
[1.4, 1.4, 0.7, 0.7, 0.7] &--11.2\\ \hline
[1.4, 1.4, 0.7, 0.5, 0.5] &--20.0\\ \hline
[1.4, 1.4, 0.7, 0.5, 0.2] &--18.5\\
\bottomrule[1.5pt]
\end{tabular}}
\label{layer_num}
\end{table}
\subsubsection{Effectiveness of Bi-directional Contextual Compression}
We visualize the forward and backward temporal contexts ($C_f^0$, $C_b^0$) of different B-frames in Fig.~\ref{ablation:tcm}. We find that the forward and backward temporal contexts can provide different temporal prediction information. For example, in the red rectangles, when the forward context of Frame 1 cannot provide accurate predictions for the horsetail, the backward context of Frame 1 supplements the information. In the green rectangles, when the backward context of Frame 1 has blurred predictions for the arm, more accurate predictions of the forward context can used. Therefore, feeding the bi-directional temporal contexts into the contextual encoder-decoder can make better use of temporal predictions. However, we find that the accuracy of temporal predictions decreases obviously when the distances between bi-directional reference frames and the current frame increase. This indicates that existing motion estimation methods for deep video compression cannot handle large motion~\cite{yang2024ucvc}. This may be the main reason why our proposed DCVC-B scheme performs worse than DCVC-DC on the UVG and MCL-JCV videos with large motion. 

\subsubsection{Effectiveness of Hierarchical Quality Structure}\label{ablation:number_layer}
To analyze why our proposed hierarchical quality structure-based training strategy can improve compression performance, we compare the frame quality and frame bitrate of the model with ($M_4$) and without ($M_3$) the training strategy. We take the \emph{BQTerrace\_1920x1080\_60} sequence and the \emph{Cactus\_1920x1080\_50} sequence of the HEVC Class B dataset as examples. As illustrated in Fig.~\ref{ablation:hqs}, we find that the hierarchical quality structure helps $M_4$ model achieve a better bit allocation within a large GOP. The $M_4$ model can allocate more bits to the reference B-frames, resulting in higher reconstruction qualities. While $M_3$ model also allocates more bits to reference B-frames, it does not improve their reconstruction qualities. For example, the 17th frame (frame index 16) of $M_3$ model is allocated more bits but its quality is the lowest. 
As listed in Table~\ref{layer_num}, we further analyze the influence of the values of hierarchical quality coefficients. We find that under the same GOP structure as Fig.~\ref{ablation:hqs}, better compression performance can be obtained by assigning larger quality coefficients to the B-frames at lower temporal layers and smaller quality coefficients to the B-frames at higher temporal layers. However, the performance improvement is not obvious when assigning a smaller coefficient (0.2) to the B-frame at Layer 5. Therefore, in this paper, we set the quality coefficients of the B-frames in Layer 1, 2, 3, 4, and 5 to [1.4, 1.4, 0.7, 0.5, 0.5], respectively.

\begin{figure}[t]
  \centering
   \includegraphics[width=\linewidth]{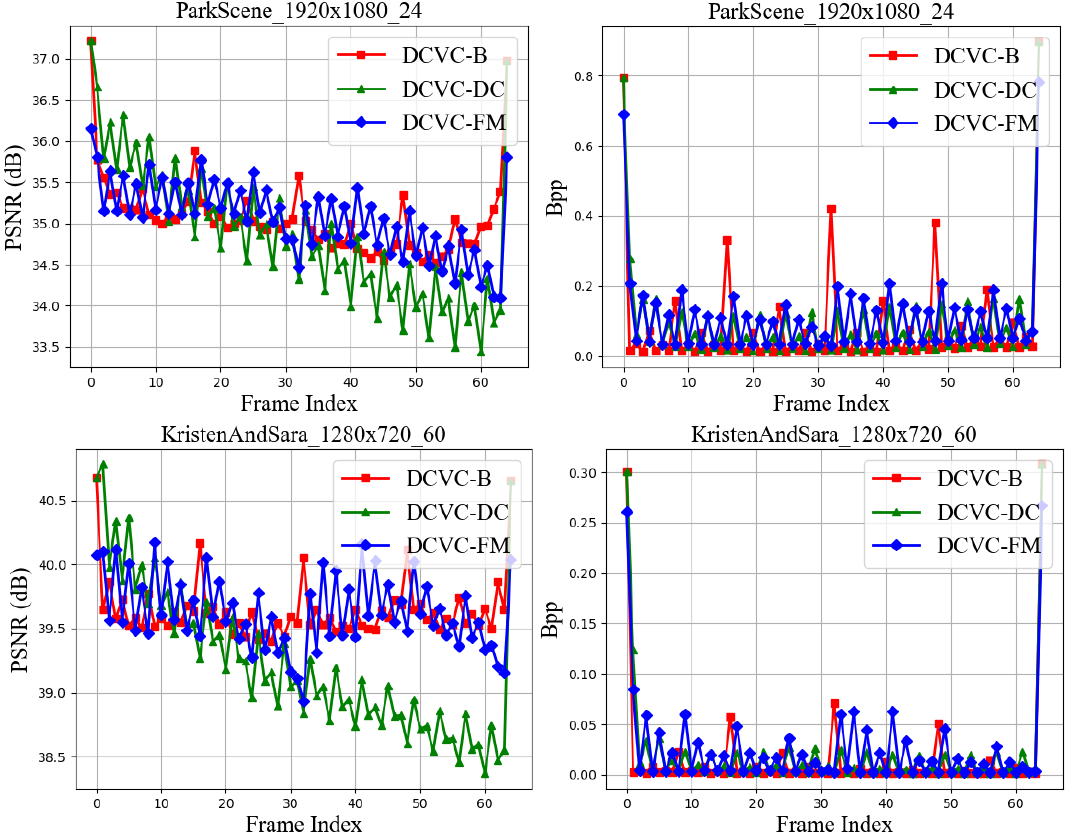}
      \caption{Frame quality and bitrate comparison between our proposed DCVC-B, DCVC-DC, and DCVC-FM when the GOP size is set to 64.}
   \label{fig:gop64}
\end{figure}
\begin{table*}[t]
 \centering
 \caption{BD-rate (\%) comparison of DCVC-DC and our DCVC-B over each testing sequence. The anchor HM-RA-GOP16.}
\begin{tabular}{l|l|c|c|c|l|l|c|c|c}
\toprule[1.5pt]
Datasets                  & Sequences          & DCVC-DC         & DCVC-FM         & Ours            & Datasets                  & Sequences  & DCVC-DC         & DCVC-FM         & Ours            \\ \hline
\multirow{5}{*}{HEVC B}   & Cactus             & --31.0          & --28.9          & \textbf{--37.2} & \multirow{30}{*}{MCL-JCV} & videoSRC01 & --17.4          & --28.4          & \textbf{--33.0} \\ \cline{2-5} \cline{7-10} 
                          & ParkScene          & --9.9           & --6.3           & \textbf{--30.8} &                           & videoSRC02 & \textbf{--16.0} & --12.0          & 0.4             \\ \cline{2-5} \cline{7-10} 
                          & Kimono1            & --29.2          & --27.8          & \textbf{--29.7} &                           & videoSRC03 & --23.4          & --20.6          & \textbf{--37.4} \\ \cline{2-5} \cline{7-10} 
                          & BQTerrace          & --12.2          & --18.4          & \textbf{--20.5} &                           & videoSRC04 & \textbf{--19.5} & --16.2          & --13.4          \\ \cline{2-5} \cline{7-10} 
                          & BasketballDrive    & --26.3          & \textbf{--26.9} & --11.8          &                           & videoSRC05 & \textbf{--30.0} & --21.4          & --11.3          \\ \cline{1-5} \cline{7-10} 
\multirow{4}{*}{HEVC C}   & BQMall             & --20.3          & \textbf{--22.7} & --14.8          &                           & videoSRC06 & --28.8          & \textbf{--46.9} & 56.1            \\ \cline{2-5} \cline{7-10} 
                          & BasketballDrill    & --25.1          & --24.0          & \textbf{--26.9} &                           & videoSRC07 & --29.9          & --20.1          & \textbf{--29.9} \\ \cline{2-5} \cline{7-10} 
                          & RaceHorses         & \textbf{--22.0} & --20.6          & --0.8           &                           & videoSRC08 & \textbf{--31.7} & --27.1          & --20.8          \\ \cline{2-5} \cline{7-10} 
                          & PartyScene         & --15.4          & --16.7          & \textbf{--23.7} &                           & videoSRC09 & --26.6          & --25.9          & \textbf{--40.1} \\ \cline{1-5} \cline{7-10} 
\multirow{4}{*}{HEVC D}   & BasketballPass     & --36.4          & --36.8          & \textbf{--38.4} &                           & videoSRC10 & \textbf{--13.6}  & 4.7             & 4.6             \\ \cline{2-5} \cline{7-10} 
                          & BQSquare           & --22.0          & --25.1          & \textbf{--42.4} &                           & videoSRC11 & \textbf{--35.5} & --25.4          & --7.6           \\ \cline{2-5} \cline{7-10} 
                          & BlowingBubbles     & --24.5          & --22.7          & \textbf{--37.3} &                           & videoSRC12 & --1.3           & --2.5           & \textbf{--31.3} \\ \cline{2-5} \cline{7-10} 
                          & RaceHorses         & \textbf{--42.4} & --37.8          & --34.9          &                           & videoSRC13 & 19.6            & 3.8             & \textbf{--18.0} \\ \cline{1-5} \cline{7-10} 
\multirow{3}{*}{HEVC E}   & KristenAndSara     & --22.8          & --27.5          & \textbf{--43.1} &                           & videoSRC14 & \textbf{--48.6} & --48.2          & --44.5          \\ \cline{2-5} \cline{7-10} 
                          & FourPeople         & --28.8          & --34.1          & \textbf{--42.4} &                           & videoSRC15 & \textbf{--30.2} & --28.4          & --28.7          \\ \cline{2-5} \cline{7-10} 
                          & Johnny             & --20.5          & --29.2          & \textbf{--43.3} &                           & videoSRC16 & --11.8          & --18.9          & \textbf{--19.4} \\ \cline{1-5} \cline{7-10} 
\multirow{6}{*}{HEVC RGB} & EBULupoCandlelight & --19.4          & --21.7          & \textbf{--28.3} &                           & videoSRC17 & --29.1          & --27.9          & \textbf{--29.6} \\ \cline{2-5} \cline{7-10} 
                          & OldTownCross       & --33.5          & --26.5          & \textbf{--35.0} &                           & videoSRC18 & \textbf{12.5}   & 19.8            & 16.4            \\ \cline{2-5} \cline{7-10} 
                          & ParkScene          & 0.3             & 15.0            & \textbf{--16.2} &                           & videoSRC19 & \textbf{--30.0} & --25.9          & --6.8           \\ \cline{2-5} \cline{7-10} 
                          & DucksAndLegs       & \textbf{--18.0} & 5.0             & --7.9           &                           & videoSRC20 & \textbf{50.2}   & 73.0            & 68.2            \\ \cline{2-5} \cline{7-10} 
                          & Kimono1            & --29.6          & --27.8          & \textbf{--29.6} &                           & videoSRC21 & \textbf{--25.7} & --22.1          & --17.1          \\ \cline{2-5} \cline{7-10} 
                          & EBURainFruits      & --6.0           & --11.3          & \textbf{--35.8} &                           & videoSRC22 & \textbf{--26.3} & --24.7          & --24.0          \\ \cline{1-5} \cline{7-10} 
\multirow{7}{*}{UVG}      & HoneyBee           & --21.7          & --16.8          & \textbf{--47.2} &                           & videoSRC23 & --37.8          & \textbf{--40.0} & --31.6          \\ \cline{2-5} \cline{7-10} 
                          & ReadySteadyGo      & \textbf{--31.2} & --27.2          & --18.1          &                           & videoSRC24 & \textbf{--33.4} & --25.9          & --30.8          \\ \cline{2-5} \cline{7-10} 
                          & Bosphorus          & --1.0           & --7.9           & \textbf{--29.5} &                           & videoSRC25 & --21.7          & --17.7          & \textbf{--23.2} \\ \cline{2-5} \cline{7-10} 
                          & YachtRide          & --26.8          & \textbf{--29.6} & --26.5          &                           & videoSRC26 & \textbf{--26.6} & --19.9          & --23.7          \\ \cline{2-5} \cline{7-10} 
                          & Jockey             & --3.0           & 23.2            & 90.9            &                           & videoSRC27 & --11.4          & --13.4          & \textbf{--23.7} \\ \cline{2-5} \cline{7-10} 
                          & ShakeNDry          & --21.7          & --23.5          & \textbf{--24.3} &                           & videoSRC28 & --20.5          & --24.9          & \textbf{--31.3} \\ \cline{2-5} \cline{7-10} 
                          & Beauty             & --61.3          & \textbf{--65.4} & --52.4          &                           & videoSRC29 & 0.2             & \textbf{--23.1} & --11.7          \\ \cline{1-5} \cline{7-10} 
-                         & -                  & -               & -               & -               &                           & videoSRC30 & --13.3          & --11.5          & \textbf{--35.8}          \\
\bottomrule[1.5pt]
\end{tabular}
\label{table:psnr_each_video}
\end{table*}
\begin{figure*}[t]
  \centering
   \includegraphics[width=\linewidth]{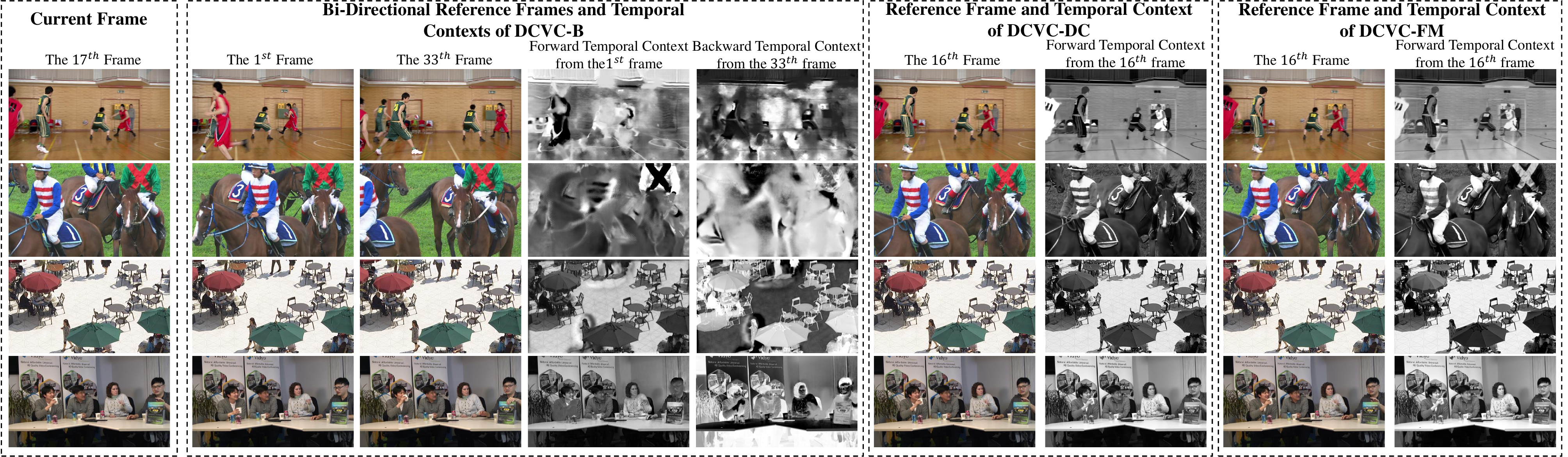}
      \caption{Visualization of the reference frames and temporal contexts of DCVC-B and DCVC-DC. The first row is the $BasketballDrive\_1920\times1080\_50$ sequence in the HEVC Class B dataset. The second row is the $RaceHorses\_832\times480\_30$ sequence in the HEVC Class C dataset. The third row is the $BQSquare\_416\times240\_60$ sequence in the HEVC Class D dataset. The fourth row is the $FourPeople\_1280\times720\_60$ sequence in the HEVC Class E dataset.}
   \label{fig:badcase}
\end{figure*}
\subsubsection{Influence of Larger GOP Size}\label{ablation:gop64}
To evaluate the capability of our DCVC-B to support a larger GOP size, we conduct an ablation study by setting its intra period and GOP size to 64. It is important to note that we did not fine-tune our model using longer training sequences. For comparison, we also set the intra period of DCVC-DC and DCVC-FM to 64. In Fig.~\ref{fig:gop64}, we compare the frame quality and bitrate of different codecs. Observing the comparison results, we find that DCVC-FM has less error propagation than DCVC-DC because it uses 32 frames for joint training and uses the feature re-fresh method. Without the need for multi-frame joint training that consumes a lot of GPU memory, our DCVC-B can rely on the bi-directional hierarchical quality structure to reduce error propagation.

\section{Limitations and Future Work}
As reported in Table~\ref{table:ip32_psnr}, our proposed DCVC-B exhibits different compression performance across various testing datasets. In this section, we aim to identify the limitations of DCVC-B by analyzing the characteristics of each dataset. We compare the BD-rate values of DCVC-DC, DCVC-FM, and DCVC-B against the anchor HM-RA-GOP16 for each video sequence in Table~\ref{table:psnr_each_video}. The comparison results reveal that DCVC-B is inferior to DCVC-DC and DCVC-FM when the testing videos have larger motion, as it struggles to obtain accurate bi-directional temporal predictions. Conversely, DCVC-B is superior to DCVC-DC and DCVC-FM when the testing videos have smaller motion, as it can leverage more useful temporal information from accurate bi-directional temporal predictions. For example, as shown in Fig.~\ref{fig:badcase}, when the motion between bi-directional reference frames is large, DCVC-B can only predict blurred bi-directional temporal contexts for sequences such as the $BasketballDrive\_1920\times1080\_50$ sequence in the HEVC Class B dataset and the $RaceHorses\_832\times480\_30$ sequence in the HEVC Class C dataset. In this case, it is difficult for DCVC-B to leverage useful temporal information from bi-directional reference frames, leading to lower compression performance. In contrast, when the motion between reference frames is small, DCVC-B can predict accurate bi-directional temporal contexts, such as the $BQSquare\_416\times240\_60$ sequence in the HEVC Class D dataset and the $FourPeople\_1280\times720\_60$ sequence in the HEVC Class E dataset. In this case, with more useful temporal information from bi-directional temporal contexts, DCVC-B can effectively reduce the conditional entropy, leading to higher compression performance. 

Regarding the limitations of DCVC-B, we identify two critical issues that need to be addressed in the future.
\begin{itemize}
    \item How to obtain more accurate bi-directional temporal contexts when the motion between reference frames is large.
    \item How to discriminately utilize bi-directional temporal contexts with prediction errors. 
\end{itemize}
By addressing these issues, we believe that higher compression performance can be achieved for deep B-frame coding schemes.
 
\section{Conclusion}\label{sec:conclusion}
In this paper, we propose a bi-directional deep contextual video compression scheme tailored for B-frames, termed DCVC-B. We focus on improving the compression performance for the B-frame coding from three aspects. Firstly, we propose a bi-directional motion difference context propagation method to reduce the motion coding costs. Secondly, we propose a bi-directional contextual compression model and a bi-directional temporal entropy model to make better use of multi-scale temporal contexts. Thirdly, we propose a hierarchical quality structure-based training strategy to achieve a better bit allocation within a large GOP. Experimental results demonstrate that, in terms of PSNR, our DCVC-B scheme significantly outperforms the reference software of H.265/HEVC under the random access configuration. On some testing datasets, DCVC-B even surpasses the reference software of H.266/VVC. Additionally, we analyze the unique challenges of deep B-frame coding and identify the limitations of our scheme. To address these limitations, we propose important future work that could further enhance the compression performance of deep B-frame coding, thereby beneficial for the community. With the gradual improvement of deep P-frame codecs in recent years, we believe that deep B-frame codecs will be better and better.
\bibliographystyle{ieeetr}
\bibliography{ref}
\begin{IEEEbiography}[{\includegraphics[width=1in,height=1.25in,clip,keepaspectratio]{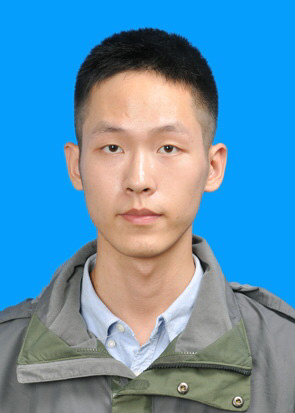}}]{Xihua Sheng} (Member, IEEE) received the B.S. degree in automation from Northeastern University, Shenyang, China, in 2019, and the Ph.D. degree in electronic engineering from University of Science and Technology of China (USTC), Hefei, Anhui, China, in 2024. 
He is currently a Postdoctoral Fellow in computer science from City University of Hong Kong. 
His research interests include image/video/point cloud coding, signal processing, and machine learning.
\end{IEEEbiography}

\begin{IEEEbiography}[{\includegraphics[width=1in,height=1.25in,clip,keepaspectratio]{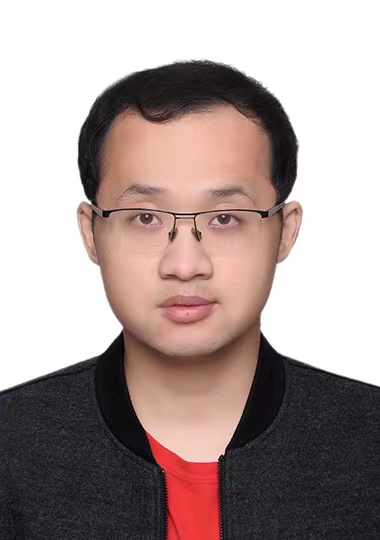}}] {Li Li} (M'17) received the B.S. and Ph.D. degrees in electronic engineering from University of Science and Technology of China (USTC), Hefei, Anhui, China, in 2011 and 2016, respectively.
He was a visiting assistant professor in University of Missouri-Kansas City from 2016 to 2020.
He joined the department of electronic engineering and information science of USTC as a research fellow in 2020 and became a professor in 2022.

His research interests include image/video/point cloud coding and processing.
He has authored or co-authored more than 80 papers in international journals and conferences. 
He has more than 20 granted patents. 
He has several technique proposals adopted by standardization groups.
He received the Multimedia Rising Star 2023.
He received the Best 10\% Paper Award at the 2016 IEEE Visual Communications and Image Processing (VCIP) and the 2019 IEEE International Conference on Image Processing (ICIP).
He serves as an associate editor for \textsc{IEEE Transactions on Circuits and Systems for Video Technology} and \textsc{IEEE Transactions on Multimedia}. 
\end{IEEEbiography}

\begin{IEEEbiography}[{\includegraphics[width=1in,height=1.25in,clip,keepaspectratio]{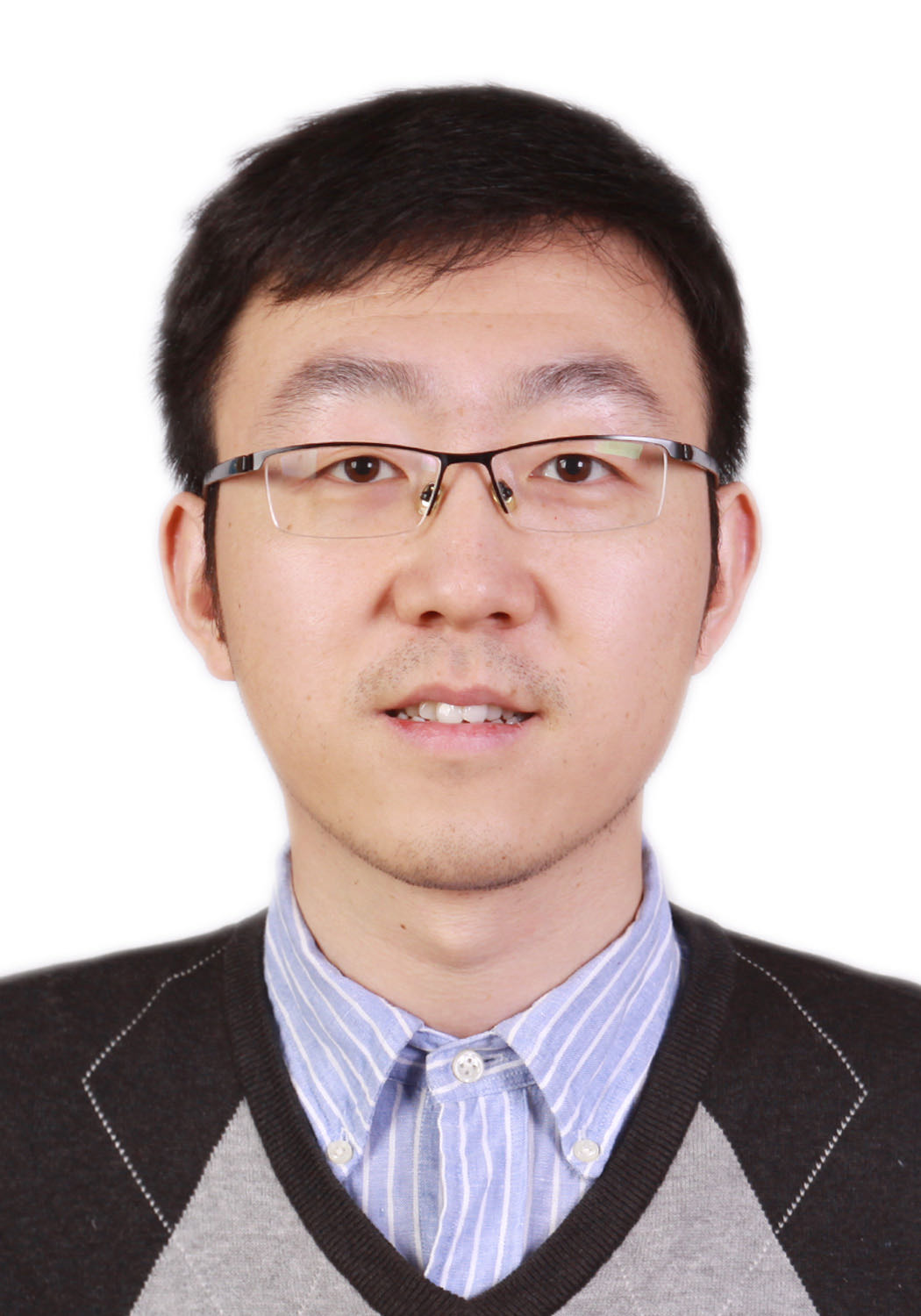}}]{Dong Liu}
(M'13--SM'19) received the B.S. and Ph.D. degrees in electrical engineering from the University of Science and Technology of China (USTC), Hefei, China, in 2004 and 2009, respectively. He was a Member of Research Staff with Nokia Research Center, Beijing, China, from 2009 to 2012. He joined USTC as a faculty member in 2012 and became a Professor in 2020.

His research interests include image and video processing, coding, analysis, and data mining.
He has authored or co-authored more than 200 papers in international journals and conferences. He has more than 30 granted patents. He has several technique proposals adopted by standardization groups.
He received the 2009 \textsc{IEEE Transactions on Circuits and Systems for Video Technology} Best Paper Award, VCIP 2016 Best 10\% Paper Award, and ISCAS 2022 Grand Challenge Top Creativity Paper Award. He and his students were winners of several technical challenges held in ISCAS 2023, ICCV 2019, ACM MM 2019, ACM MM 2018, ECCV 2018, CVPR 2018, and ICME 2016. He is a Senior Member of CCF and CSIG, and an elected member of MSA-TC of IEEE CAS Society. He serves or had served as the Chair of IEEE 1857.11 Standard Working Subgroup (also known as Future Video Coding Study Group), an Associate Editor for \textsc{IEEE Transactions on Image Processing}, a Guest Editor for \textsc{IEEE Transactions on Circuits and Systems for Video Technology}, an Organizing Committee member for VCIP 2022, ChinaMM 2022, ICME 2021, etc.
\end{IEEEbiography}

\begin{IEEEbiography}[{\includegraphics[width=1in,height=1.25in,clip,keepaspectratio]{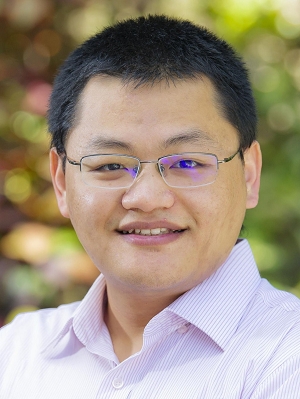}}] {Shiqi Wang} (Senior Member, IEEE) received the PhD degree in computer application technology from Peking University, in 2014. He is currently an associate professor with the Department of Computer Science, City University of Hong Kong. He has proposed more than 70 technical proposals to ISO/MPEG, ITUT, and AVS standards. He authored or coauthored more than 300 refereed journal articles/conference papers, including more than 100 IEEE Transactions. His research interests include semantic and visual communication, AI generated content management, machine learning, information forensics and security, and image/video quality assessment. He received the Best Paper Award from IEEE VCIP 2019, ICME 2019, IEEE Multimedia 2018, and PCM 2017. His coauthored article received the Best Student Paper Award in the IEEE ICIP 2018. He served or serves as an associate editor for IEEE Transactions on Circuits and Systems for Video Technology, IEEE Transactions on Multimedia, IEEE Transactions on Image Processing, and IEEE Transactions on Cybernetics.
\end{IEEEbiography}

\end{document}